\documentclass[journal]{IEEEtran}

\ifCLASSINFOpdf

\else

\fi

\usepackage{amsmath}

\usepackage{algorithmic}

\usepackage{array}

\usepackage{url}

\usepackage{multirow}

\usepackage{graphics}
\usepackage{color}
\usepackage{cite}
\usepackage{times,epsfig}
\usepackage{epstopdf}
\usepackage{subfigure}
\usepackage{amssymb}
\usepackage{url}
\usepackage{diagbox}
\begin{document}
\title{Image and Video Compression with Neural Networks: A Review}

\author{Siwei~Ma,~\IEEEmembership{Member,~IEEE,}
        Xinfeng~Zhang,~\IEEEmembership{Member,~IEEE,}
        Chuanmin~Jia,
        Zhenghui~Zhao,
        Shiqi~Wang,~\IEEEmembership{Member,~IEEE,}
        and~Shanshe~Wang
\thanks{Manuscript received Oct. 10, 2018; revised Feb. 2, 2019 and Mar. 10,  2019;  accepted  Apr. 1, 2019. This work was supported by the National Natural Science Foundation (61632001,61571017), and was also partially supported by Hong Kong RGC Early Career Scheme under Grant 9048122 (CityU 21211018) and Peng Cheng Lab. (\textit{Corresponding  author: Prof. Siwei Ma})}
\thanks{Siwei Ma, Chuanmin Jia, Zhenghui Zhao and Shanshe Wang are with the Institute of Digital Media, School of Electronic Engineering and Computer Science, Peking University, Beijing 100871, China, and are also with the Peng Cheng Lab, Shenzhen, China (e-mail: \{swma, cmjia, zhzhao, sswang\}@pku.edu.cn).}
\thanks{Xinfeng Zhang is with the Department of Computer Science, University of Chinese Academy of Sciences, Beijing 100049, China (e-mail: zhangxinf07@gmail.com).}
\thanks{Shiqi Wang is with the Department of Computer Science, City University of Hong Kong, Kowloon, Hong Kong (e-mail: shiqwang@cityu.edu.hk).}
}

\markboth{IEEE Transactions on Circuits and Systems for Video Technology}%
{Shell \MakeLowercase{\textit{et al.}}: Bare Demo of IEEEtran.cls for IEEE Journals}

\maketitle

\begin{abstract}
In recent years, the image and video coding technologies have advanced by leaps and bounds. However, due to the popularization of image and video acquisition devices, the growth rate of image and video data is far beyond the improvement of the compression ratio. In particular, it has been widely recognized that there are increasing challenges of pursuing further coding performance improvement within the traditional hybrid coding framework. Deep convolution neural network (CNN) which makes the neural network resurge in recent years and has achieved great success in both artificial intelligent and signal processing fields, also provides a novel and promising solution for image and video compression. In this paper, we provide a systematic, comprehensive and up-to-date review of neural network based image and video compression techniques. The evolution and development of neural network based compression methodologies are introduced for images and video respectively. More specifically, the cutting-edge video coding techniques by leveraging deep learning and HEVC framework are presented and discussed, which promote the state-of-the-art video coding performance substantially. Moreover, the end-to-end image and video coding frameworks based on neural networks are also reviewed, revealing interesting explorations on next generation image and video coding frameworks/standards. The most significant research works on the image and video coding related topics using neural networks are highlighted, and future trends are also envisioned. In particular, the joint compression on semantic and visual information is tentatively explored to formulate high efficiency signal representation structure for both human vision and machine vision, which are the two dominant signal receptor in the age of artificial intelligence.
\end{abstract}


\IEEEpeerreviewmaketitle

\section{Introduction}

\IEEEPARstart{I}{mage} and video compression plays an important role in providing high quality image/video services under the limited capabilities of transmission networks and storage. The redundancies within images and videos are fundamentally important for image and video compression, including spatial redundancy, visual redundancy and statistical redundancy. Besides, the temporal redundancy existing in video sequences enables the video compression to achieve higher compression ratio compared with image compression.

For image compression, the early methods mainly realize compression by directly utilizing the entropy coding to reduce statistical redundancy within the image, such as Huffman coding \cite{huffman1952method}, Golomb code \cite{golomb1966run} and arithmetic coding \cite{witten1987arithmetic}. In later 1960s, transform coding was proposed for image compression by encoding the spatial frequencies, including Fourier transform \cite{andrews1968fourier} and Hadamard transform \cite{pratt1969hadamard}. In 1974, Ahmed \textit{et al.} proposed Discrete Cosine Transform (DCT) for image coding \cite{ahmed1974discrete}, which can compact image energy in the low frequency domain such that compression in the frequency domain becomes much more efficient.

Besides reducing statistical redundancy by entropy coding and transform techniques, the prediction and quantization techniques are further proposed to reduce spatial redundancy and visual redundancy in images. The most popular image compression standard, JPEG, is a successful image compression system by integrating its preceding coding techniques. It first divides image into blocks and then transforms blocks into the DCT domain. For each block, the differential pulse code modulation (DPCM) \cite{harrison1952experiments} is applied to its DC components, such that the prediction residuals of DC components between neighboring DCT blocks are compressed instead of compressing the DC value directly. To reduce the visual redundancy, a special quantization table is designed to well preserve low-frequency information and discard more high-frequency (noise-like) details as humans are less sensitive to the information loss in high frequency parts \cite{wallace1990overview}. Another well-known still image compression standard, JPEG 2000\cite{christopoulos2000jpeg2000}, applies the 2D wavelet transform instead of DCT to represent images in a compact form, and utilizes an efficient arithmetic coding method, EBCOT \cite{taubman2000high}, to reduce the statistical redundancy existing in wavelet coefficients.

For video coding, temporal redundancy, which could be removed by inter-frame prediction, becomes the dominant one due to the high correlation between successive frames captured in a very short time interval. To acquire inter-prediction efficiently, the block based motion prediction was proposed in 1970s \cite{taki1974interframe}. In 1979, Netravali and Stuller proposed motion compensation transform framework \cite{netravali1979motion}, which is well known as the hybrid prediction/transform coder nowadays. Reader provided an introduction to the historical development of the first generation methods \cite{reader2002history}.

\begin{figure}[t]
\centering
\begin{minipage}[b]{0.235\textwidth}
\centering
\subfigure[]{\label{fig:angularillustration}\includegraphics[width=4.0cm]{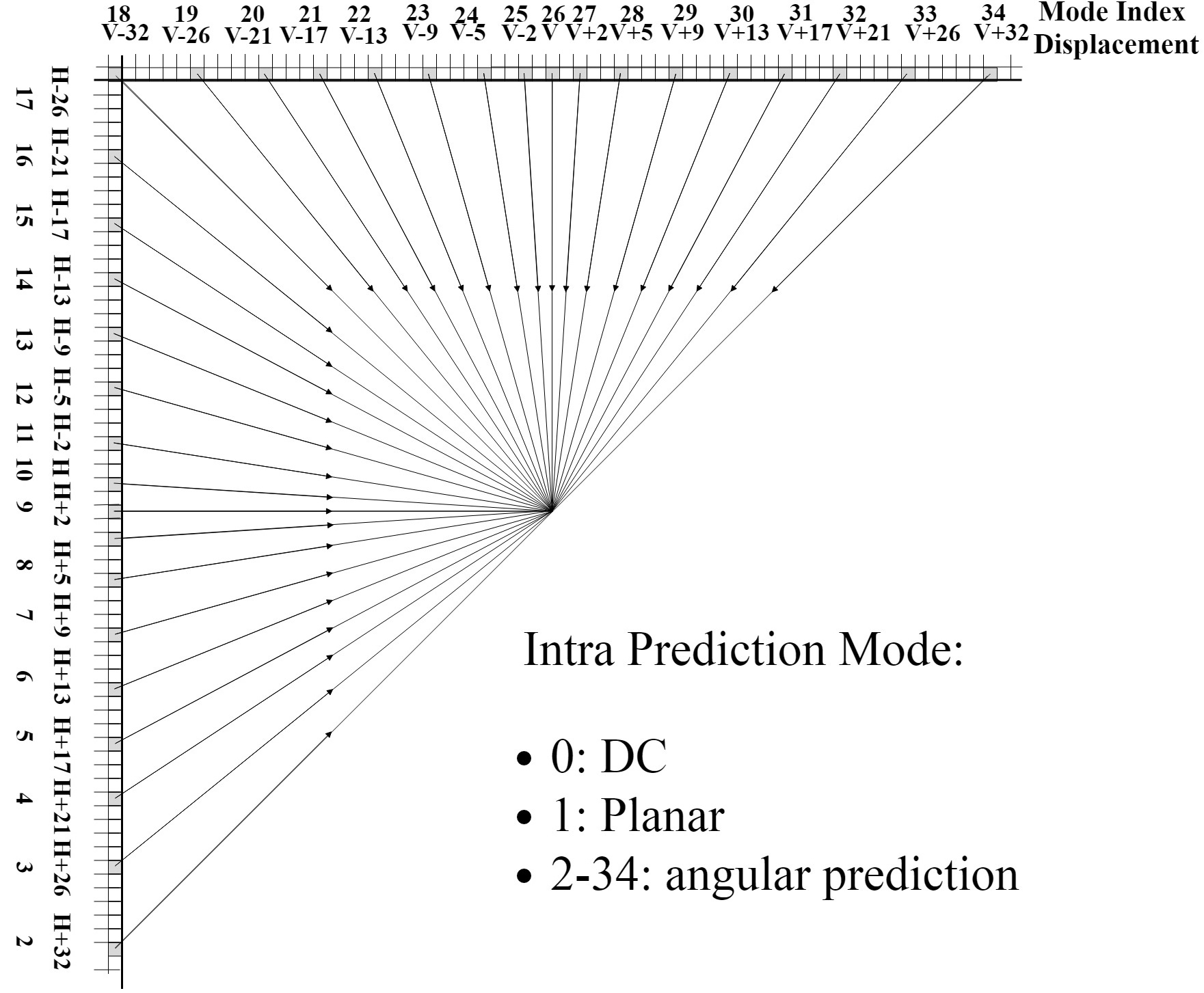} }
\end{minipage}
\hspace{0.05cm}
\begin{minipage}[b]{0.235\textwidth}
\centering
\subfigure[]{\label{fig:intraexample}\includegraphics[width=3.0cm]{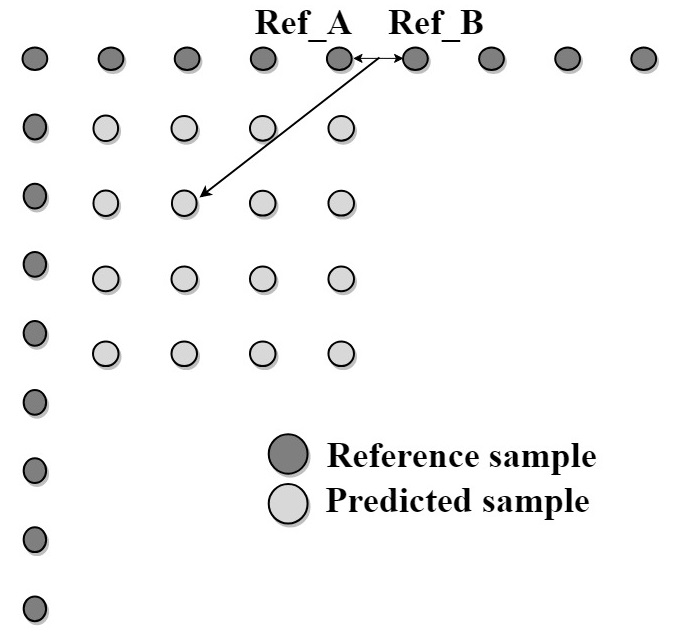} }
\end{minipage}

\caption{Illustration of HEVC intra prediction. (a) different intra modes; (b) angular prediction instance.}
\label{Fig:hevc-intra}
\end{figure}

After several decades of development, the hybrid prediction/transform coding methods have achieved great success. Many coding standards have been developed and widely used in various applications, such as MPEG-1/2/4, H.261/2/3 and H.264/AVC\cite{wiegand2003overview}, as well as AVS (Audio and Video coding Standard in China) \cite{avs.org} and HEVC \cite{gary2013overview}. Taking the latest video coding standard, HEVC, as an example, it utilized neighboring reconstructed pixels to predict the current coding block, with 33 angular intra prediction modes, the DC mode and the planar mode, as shown in Fig.~\ref{Fig:hevc-intra}. For inter-frame coding, HEVC improves the coding performance by further refining its predecessor, H.264/AVC, from multiple perspectives, e.g., increasing the diversity of the PU division, utilizing more interpolation filter taps for sub-sample motion compensation \cite{lv2012comparison} and refining the side information coding including more most probable modes (MPMs) for intra mode coding \cite{lainema2012intra}, advanced motion vector prediction (AMVP) and merge mode for motion vector predictor coding \cite{lin2013motion}. Another new video coding tool in state-of-the-art video coding framework is loop filtering, and many loop filters \cite{naccari2012adaptive,zhang2017low,ma2016nonlocal,tsai2013adaptive,zhang2012adaptive,zhang2017high} have been proposed since 2000. Herein, the deblock filtering\cite{list2003adaptive,norkin2012hevc} and sample adaptive offset (SAO) \cite{fu2012sample} has been adopted into HEVC. However, the refinement strategies for traditional hybrid video coding framework based on image and video local correlations are more and more difficult for further coding efficiency improvement.

Recently, neural networks, especially the convolution neural networks (CNN), have achieved significant success in many fields including the image/video understanding, processing and compression etc. \textbf{A CNN is usually comprised of one or more convolutional layers. In particular, some tasks also append several fully connected layers after the convolution layers}. The parameters in these layers can be well trained based on massive image and video samples labelled for specific tasks in an end-to-end strategy. The trained CNN can be well applied to solve classification, recognition and prediction tasks on test data with highly efficient adaptability. The quality of the prediction signals generated by CNN has surpassed that of the rule-based predictors. Moreover, the CNN can be interpreted as feature extractors to transform the image and video into feature space with compact representation, which is beneficial for image and video compression. Based on these excellent characteristics of CNN, it has also been recognized as a promising solution for compression task. Therefore, to well understand the existing development of CNN on image and video compression, this paper provides a detailed review on image and video compression using neural network.

\begin{figure}[t]
\center{
\includegraphics[width=0.43\textwidth]{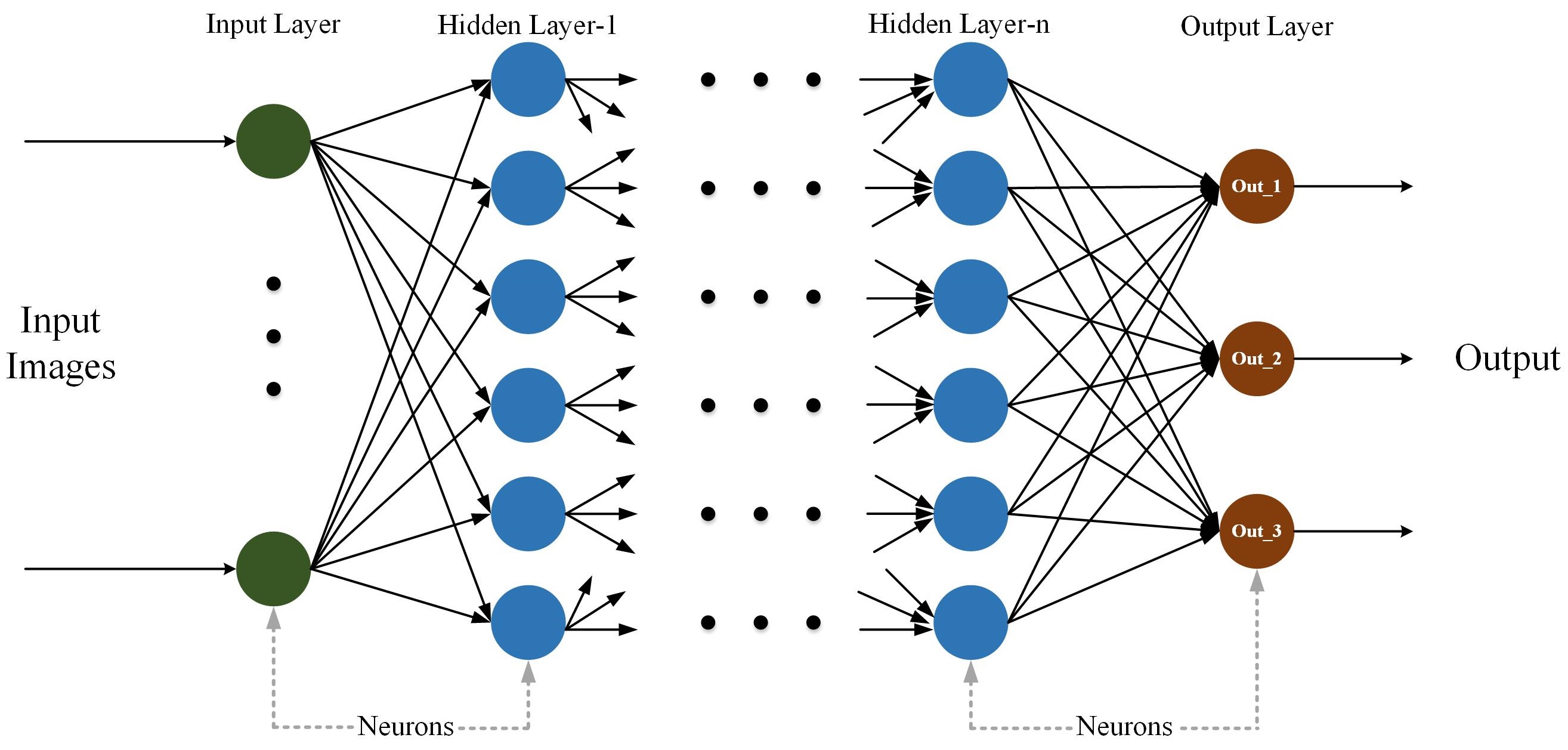}}
\caption{Illustration of the neural network architecture.}
\label{Fig:SimpleNN}
\end{figure}

Due to the vast scope of this review, we divide the main body of the paper into four parts for clearer presentation. In section \ref{neuralnetwork}, we introduce the basic concept for neural network and image/video compression. Section \ref{CNNImage} provides a detailed review on the development of neural network based image compression techniques. In section \ref{CNNVideo}, we review the techniques of neural network based video compression. In section \ref{codingopt}, we revisit the neural network based optimization techniques for image and video compression. The further rationale in section \ref{CNNImage} mainly follows the timeline of network development to introduce the neural network based image compression based on representative network architectures. In section \ref{CNNVideo}, we mainly focus on the CNN based video coding techniques imbedded in the state-of-the-art hybrid video coding framework, HEVC, and also introduce some new video coding frameworks based CNN. Finally, section \ref{conclusion} prospects the important challenges in deep learning based image/video compression and concludes the paper.

\begin{figure*}[t]
\centering
\begin{minipage}[b]{0.45\textwidth}
\centering
\subfigure[]{\label{fig:imagecompression}\includegraphics[width=6.9cm]{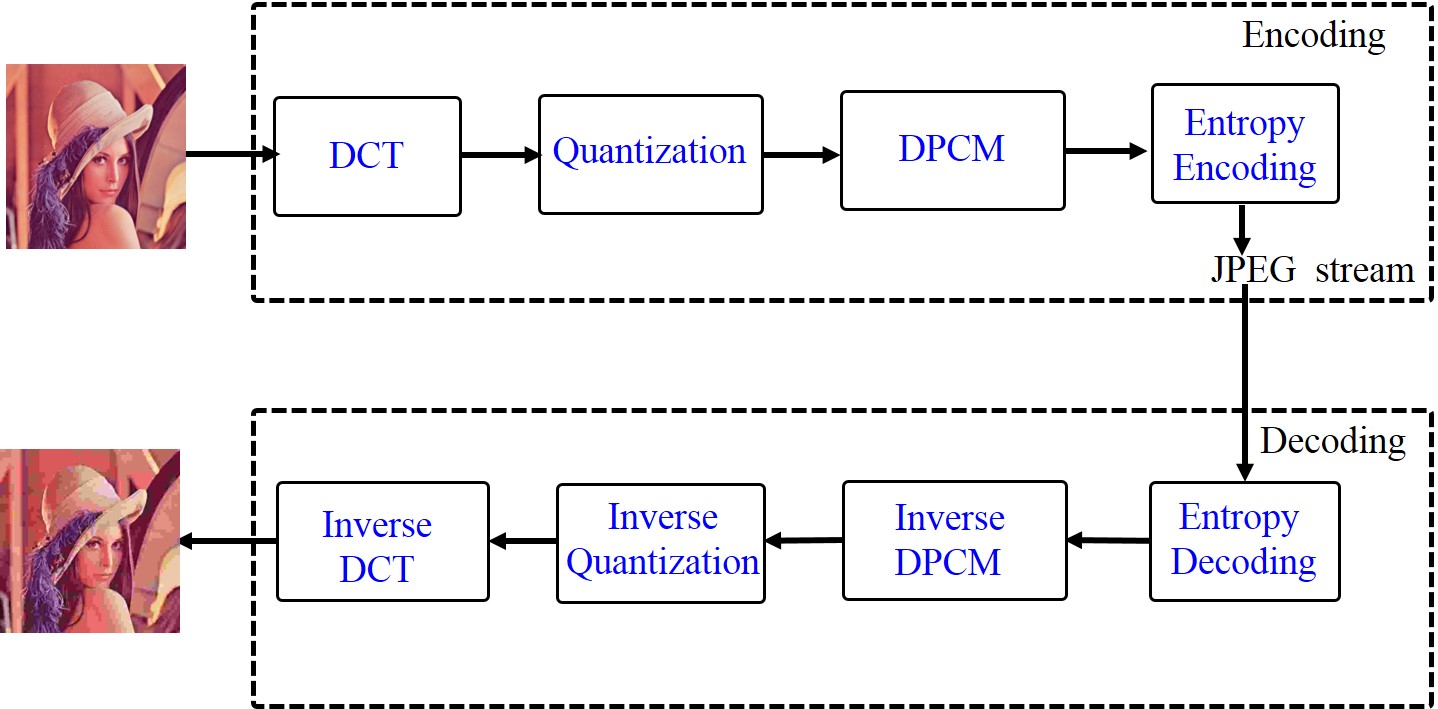} }
\end{minipage}
\hspace{0.5cm}
\begin{minipage}[b]{0.45\textwidth}
\centering
\subfigure[]{\label{fig:videocompression}\includegraphics[width=6.9cm]{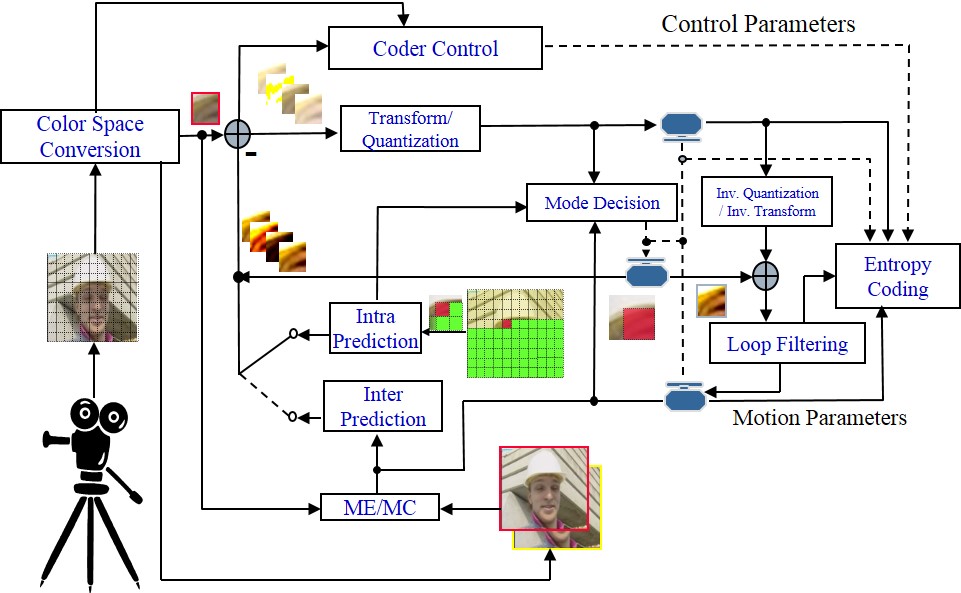} }
\end{minipage}

\caption{Image and video compression framework (a) JPEG compression, (b) hybrid video compression.}
\label{fig:image-video-coding}
\end{figure*}

\begin{figure*}[t]
\center{
\includegraphics[width=0.88\textwidth]{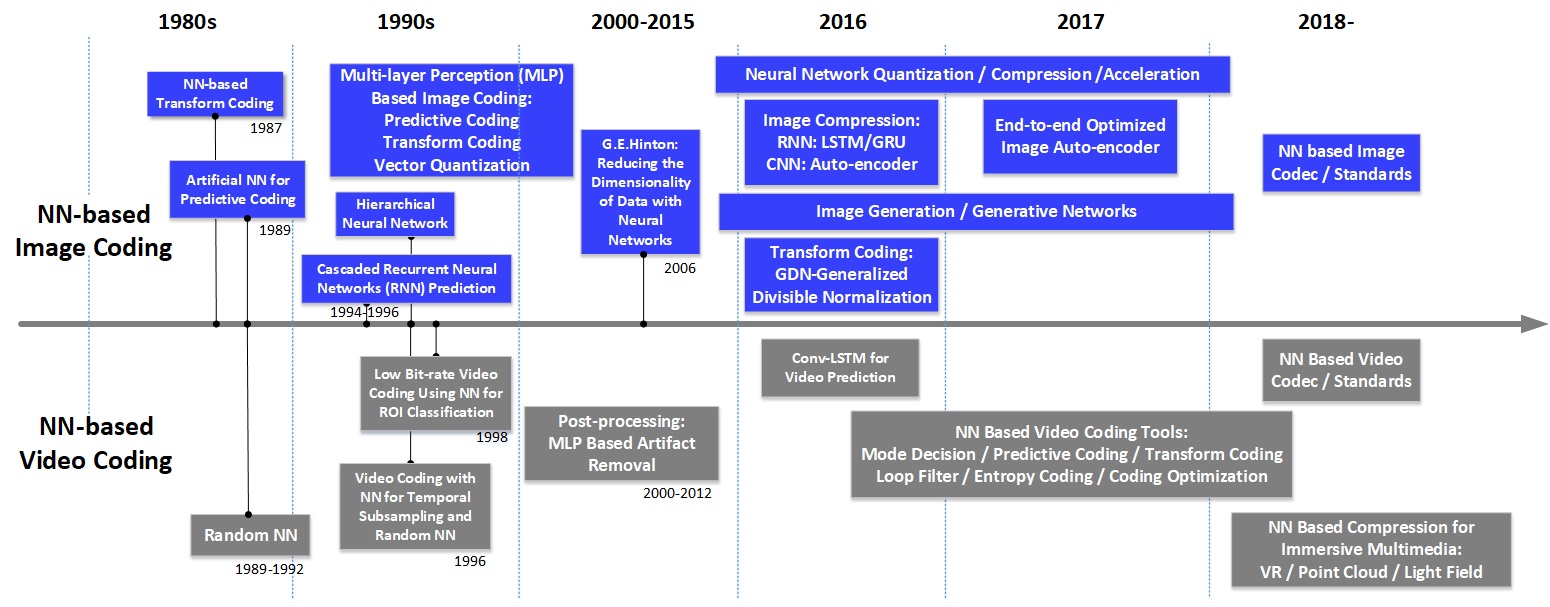}}
\caption{The technical roadmap of neural network based compression algorithms.}
\label{Fig:roadmap}
\end{figure*}

\section{Introduction of Neural Network and Image/Video Compression}\label{neuralnetwork}
In this section, we firstly revisit the basic concepts and development history of neural networks briefly. Subsequently, we introduce the frameworks and basic technique development for block based image coding and hybrid video coding framework.

\subsection{Neural Network}
With the interdisciplinary research of neuroscience and mathematics, the neural network (NN) was invented, which has shown strong abilities in the context of non-linear transform and classification. Intuitively, the network consists of multiple layers of simple processing units called neuron (perceptron), which interacts with each other via weighted connections. The neurons get activated through weighted connections from previously activated neurons.
To achieve non-linearity, the activation functions are always applied for all the intermediate layers~\cite{hinton1987learning}. A simple neural network architecture is shown in Fig.~\ref{Fig:SimpleNN}, which consists of one input layer, one output layer and multiple hidden layers, each of which contains various number of neurons.

The learning procedure of simple perceptron has been proposed and analyzed in 1960s \cite{rosenblatt1962principles}. During the 1970s and 1980s, backpropagation procedure ~\cite{werbos1974new,rumelhart1986learning} inspired by the chain rule for derivatives of the training objectives was proposed to solve the training problem of the multi-layer perceptron (MLP). Then, the multi-layer architectures are mostly trained by stochastic gradient descent with backpropagation procedure although it is computationally intensive and suffers from bad local minima. However, the dense connections between the adjacent layers in neural networks make the amount of model parameters increase quadratically and prohibit the development of neural networks in computational efficiency. With the introduction of parameter-sharing for MLP 1990~\cite{le1990handwritten}, a more light-weighted version of neural network called convolutional neural network was proposed and applied in the documents recognition, which makes the large scale neural network training possible.

\subsection{Image and Video Compression}
Among the various coding frameworks, the core techniques in image and video compression are transform and prediction. JPEG \cite{wallace1991jpeg} is the most popular image compression standard, which consists of the basic transform/prediction modules as shown in Fig.~\ref{fig:imagecompression}. In JPEG, the input image is partitioned into $8\times 8$ non-overlapped blocks, each of which is transformed into the frequency domain using block-DCT (BDCT). For each transformed block, the DCT coefficients are then compressed into a binary stream via quantization and entropy coding. For video compression, most of popular video coding standards adopt the transform-prediction based hybrid video coding framework as shown in Fig.~\ref{fig:videocompression}, e.g., MPEG-2, H.264/AVC and HEVC. Different from JPEG, HEVC utilizes more intra prediction modes from neighboring reconstructed blocks in spatial domain instead of DC prediction, as shown in Fig.~\ref{Fig:hevc-intra}. Besides intra prediction, more coding gains of video compression come from the high efficient inter prediction, which utilizes motion estimation to find the most similar blocks as prediction for the to-be-coded block. Moreover, HEVC adopts two loop filters, i.e., deblocking filter and SAO, to reduce the compression artifacts sequentially.


In the above block based image and video coding standards, the compression is usually block-dependent and must be performed block by block sequentially, which limits the compression parallelism using parallel computation platform, e.g. GPU. Moreover, the independent optimization strategy for each individual coding tool also limits the compression performance improvement compared with end-to-end optimization compression. In essence, there is another technological development trajectory based on the neural network techniques for image and video compression as summarized in Fig.~\ref{Fig:roadmap}. With the resurgence of neural network, the marriage of traditional image/video compression and CNN further advances their progress. In the following sections, we will introduce the development of neural network based image/video compression and related representative techniques.

\section{Progress of Neural Network Based Image Compression}\label{CNNImage}
In this section, we introduce the image compression using machine learning methods especially from neural network perspective, which mainly originated from late 1980s~\cite{chua1988neural}.
This section is organized according to the historical development of neural network techniques, mainly including the  Multi-layer Perceptron (MLP), Random Neural Network, Convolutional Neural Network (CNN) and Recurrent Neural Networks (RNN). In the final subsection, we will introduce the recent development of the image coding techniques using generative adversarial networks (GAN).


\subsection{Multi-layer Perceptron based Image Coding}
MLP~\cite{gardner1998artificial} consists of an input layer of neurons (or nodes, units), several hidden layers of neurons, and a final layer of output neurons. The output $h_{i}$ of each neuron ${i}$ within the MLP is denoted as,
\begin{equation} \label{Eq:ann-neuron}
h_{i}=\sigma(\Sigma_{j=1}^{N}w_{ij}x_{j}+c_{i}),
\end{equation}
where $\sigma(\cdot)$ is the activation function, $c_{i}$ denotes the bias-term of linear transform and the $w_{ij}$ indicates the adjustable parameter, \textit{weight}, which represents the connection between layers. The theoretical analysis has shown that the MLP constructed with over one hidden layer can approximate any continuous computable function to an arbitrary precision~\cite{schalkoff1997artificial}. This property provides the evidence for the scenarios such as dimension reduction and data compression. The initiative of using MLP for image compression is to design unitary transforms for the whole spatial data.

\begin{figure}[t]
\center{
\includegraphics[width=0.9\columnwidth]{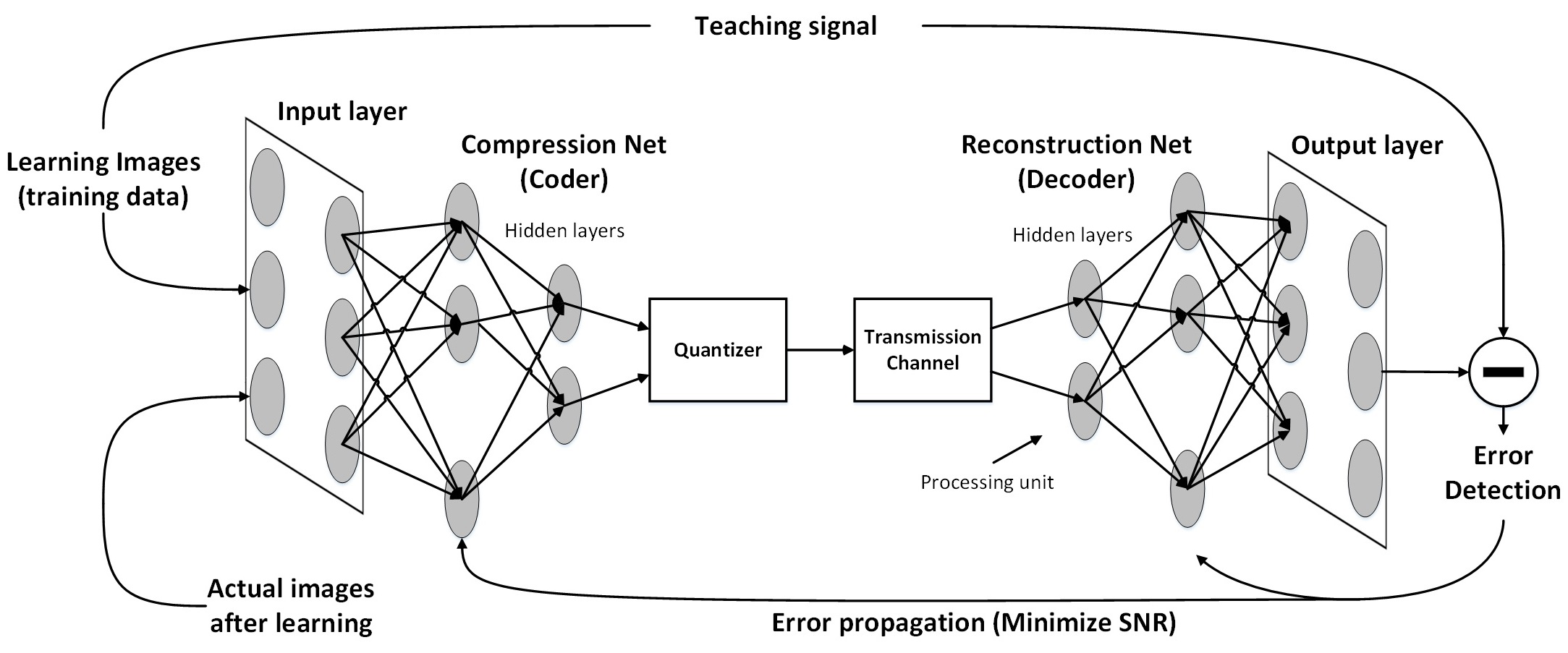}}
\caption{The Neural Network based image codec~\cite{sonehara1989image}.}
\label{Fig:neuralcodec}
\end{figure}

In 1988, Chua and Lin proposed an end-to-end image compression framework by leveraging high parallelism and the powerful compact representation of neural network \cite{chua1988neural}, which may be useful as a model of the human brain-like coding functions. They formulated the traditional image compression steps, i.e., the unitary transform of spatial domain image data, the quantization of transform coefficients and binary coding of quantized coefficients, as an integrated optimization problem to minimize the following cost function,
\begin{equation}
e_{k,l} = \|\mathbf{X} - \hat{y}_{kl}\mathbf{u}_k^T\mathbf{u}_l\|^2
\label{Eq:Objectivefunction}
\end{equation}
\begin{equation}
\hat{y}_{kl} = (s_12^{-1} + s_22^{-2} + \cdots + s_{kl}2^{-b_{kl}} ) - m_{k,l},
\label{Eq:quantizedC}
\end{equation}
where $\hat{y}_{kl}\in [0,1]$ is the reconstructed transform coefficients, $\mathbf{u}_k$ is the orthogonal transform kernel and $\{s_1,s_2,\cdots,s_{b_{kl}}\}$ are the binary codes representing the quantization level for $\hat{y}_{kl}$. Then, the authors utilized a decomposition/decision neural network to solve the optimization problem in Eqn.~(\ref{Eq:Objectivefunction}) to find the optimal binary code combination, which is the output of the compressed bitstream. In 1989, the fully connected neural network with 16 hidden units was trained to compress each $8\times8$ patch of an image using back propagation~\cite{munro1989image}. However, this strategy fixed the neural network parameters for specific number of binary codes, which is difficult to adapt to variable compression ratio in the optimal state.

Sonehara \textit{et al.} proposed to train a dimension reduction neural network to compress the input image, and took the quantization and entropy coding as individual modules \cite{sonehara1989image}. Fig.~\ref{Fig:neuralcodec} shows the architecture of the dimension reduction neural network, where the auto-encoder bottleneck structure is deployed. In particular, the number of neurons in the bottleneck layer is smaller than the number of neurons in the input and output layers so as to reduce the dimension of data. To speed up the learning process, the input image is divided into blocks, which are fed to different sub-neural networks in parallel. The design using multiple sub-neural networks requires the input image to be strictly similar to the learned ones due to different sub-neural networks being responsible for texture-specific structures. Therefore, the generalization of this model is limited, specifically, a loss of up to 10dB in SNR for unlearned images is reported. To obtain better performance and generalization capability, Sicuranza \textit{et al.} trained a unique small neural network by feeding the image blocks sequentially, the loss of which is only about 1dB in SNR from unlearned images to learned images \cite{sicuranza1990artificial}.

\begin{figure}[t]
\centering
\begin{minipage}[b]{0.1\textwidth}
\centering
\subfigure[]{\label{fig:annpredict1}\includegraphics[width=1.6cm]{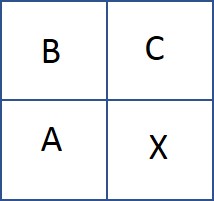} }
\end{minipage}
\begin{minipage}[b]{0.3\textwidth}
\centering
\subfigure[]{\label{fig:annpredict2}\includegraphics[width=5.2cm]{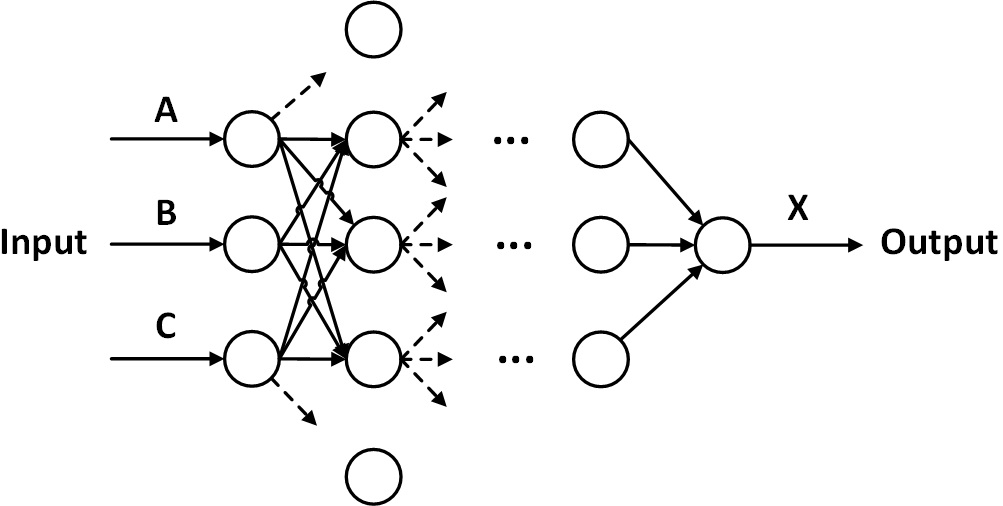} }
\end{minipage}

\caption{Illustration of MLP-based predictive coding for image compression~\cite{dianat1991non}.}
\label{Fig:ann-spatial-pred}
\end{figure}

However, the adaptivity of above mentioned algorithms is determined by manually-setting different number of hidden neurons rather than bringing networks with more layers and complex connections, which may restrict the power of MLP in terms of compression performance~\cite{dony1995neural}. To tackle this problem, MLP-based predictive image coding algorithm~\cite{dianat1991non} was investigated by exploiting the spatial context information. Specifically, the spatial information to the left and above (points $A$, $B$ and $C$, each small block corresponds to one pixel in Fig.~\ref{fig:annpredict1}) was adopted to generate the non-linear predictor of the bottom-right pixel $X$ in Fig.~\ref{fig:annpredict1}. There are three input nodes, 30 hidden nodes and one output node for the MLP predictor as shown in Fig.~\ref{fig:annpredict2}, and the MLP model is trained by utilizing the back propagation algorithm \cite{rumelhart1986learning} to minimize the mean square errors between original and predicted signals. Based on their experiments, the MLP based non-linear predictor achieves an improvement on error entropy from 4.7 bits per pixel (bpp) to 3.9 bpp compared with linear predictor.

To further improve the prediction accuracy, Manikopoulos utilized a high-order prediction model as in Eqn.(\ref{Eq:ar-model-pred}) for a generalized auto-regressive (AR) model, which can well handle the sharply defined structures such as edges and contours in images \cite{manikopoulos1992neural}.
\begin{multline}\label{Eq:ar-model-pred}
x_{n}=\Sigma_{i}w_{i}x(n-i) + \Sigma_{i}\Sigma_{j}w_{ij}x(n-i)x(n-j) + \\
\Sigma_{i}\Sigma_{j}\Sigma_{k}w_{ijk}x(n-i)x(n-j)x(n-k) + ... + \epsilon_{n},
\end{multline}
where $\epsilon_{n}$ is a sequence of zero-mean i.i.d. random variables.
In 1996, the hierarchical neural network with its Nested Training Algorithm (NTA) was proposed for MLP based image compression~\cite{namphol1996image}, which considerably reduced the training time.
The interested reader can refer to \cite{daugman1988complete,abbas1993neural} to known more MLP based image compression techniques, which improve the compression efficiency by designing different connection strucures.

\subsection{Random Neural Network based Image Coding}
A new class of random neural network \cite{gelenbe1989random} was introduced in the 1989. Random neural network performs differently from the above mentioned MLP based methods in which signals are in spatial domain and optimized by the gradient backpropagation method. The signals in random neural network are transmitted in the form of spikes of unit amplitude. The communication between these neurons is modeled as a Poisson process where positive signals represent excitatory signals and negative signals represent inhibition. Some theoretical results were presented to analyze the behavior of random neural network in \cite{gelenbe1989random}. A ``backpropagation'' type training method is adopted to update the parameters, which requires the solution of $n$ linear and $n$ non-linear equations each time with a new input-output pair.

Some researchers considered the combination of the random neural network and image compression, and presented some meaningful results. Gelenbe \emph{et al.} first applied the random neural network in the image compression task~\cite{gelenbe1994random}. The architecture adopts a feedforward encoder/decoder random neural network with one intermediate layer. In particular, the first layer takes an image as the input, the last layer outputs a reconstructed image and the intermediate layer products compressed bits. Cramer \textit{et al.} further extend the work in \cite{gelenbe1994random} by designing a adaptive still block-by-block random neural network compression/decompression \cite{cramer1996video}. There are multiple distinct neural compression networks $C_1$, ... ,$C_L$ which are designed to achieve different compression levels. Each of these networks compresses the block in parallel, and the choice of the networks is select according to the quality of decompressed results. Hai further improved the compression performance by integrating the random neural network into the wavelet domain of images \cite{hai2001video}.


\subsection{Convolutional Neural Network based Coding}


Recently, CNN outperforms the traditional algorithms by a huge margin in high-level computer vision tasks such as the image classification, object detection~\cite{lecun2015deep}. Even for many low-level computer vision tasks, it also achieves very impressive performance, e.g., super-resolution and compression artifact reduction. CNN adopts the convolution operation to characterize the correlation between neighboring pixels, and the cascaded convolution operations well conform the hierarchical statistical properties of natural images. In addition, the local receptive fields and shared weights introduced by the convolution operations also decrease trainable parameters of CNN, which significantly reduce the risk of the over-fitting problem. Inspired by powerful representation of CNN for images, many works have been carried out to explore the feasibility of CNN-based lossy image compression.

However, it is difficult to straightforwardly incorporate the CNN model into end-to-end image compression. Generally speaking, CNN training depends on the back-propagation and stochastic gradient descent algorithm which demand the almost-everywhere differentiability of the loss function with respect to the trainable parameters such as the convolution weights and biases. Due to the quantization module in image compression, it produces zero gradients almost everywhere which stops the parameters updating in the CNN. In addition, the classical rate-distortion optimization is difficult to be applied to CNN based compression framework. This is because the end-to-end training for CNN needs a differentiable loss function, but the rate must be calculated based on the population distribution of the whole quantized bins, which is usually non-differentiable with respect to arguments in CNN.

\begin{figure}[t]
\center{
\includegraphics[width=0.43\textwidth]{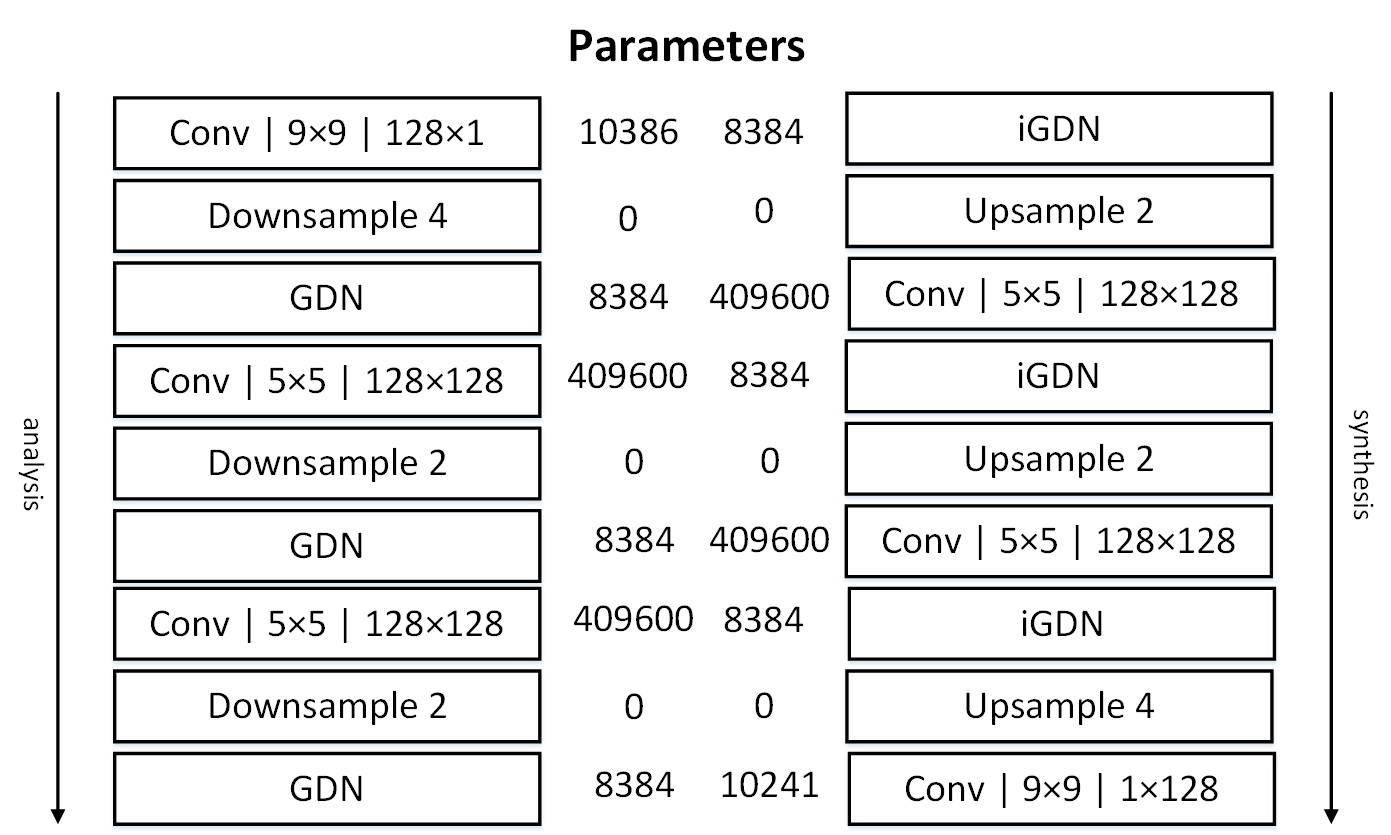}}
\caption{The parameterized architecture of the CNN based end-to-end image compression proposed in \cite{balle2016end}.}
\label{Fig:AutoEncoderImage}
\end{figure}



Ball\'{e} \emph{et al.} first introduced an end-to-end optimized CNN framework for image compression under the scalar quantization assumption in 2016 \cite{balle2016end,balle2016endpcs}. The framework is illustrated in Fig.~\ref{Fig:AutoEncoderImage}, which consists of two modules, \textit{i.e.,} analysis and synthesis transforms for encoder and decoder. In analysis transform, there are three stages of convolution, subsampling, and divisive normalization. Each stage starts with an affine convolution:
\begin{equation}
v_i^{(k)}(m,n) = \sum(h_{k,ij}\ast u_j^{(k)})(m,n) + c_{k,i},
\label{convolution}
\end{equation}
where $u_j^{(k)}$ is the $j^{th}$ input channel of the $k^{th}$ stage at spatial location $(m,n)$, $\ast$ denotes 2D convolution operation and $h_{k,ij}$ represents the convolution parameter. $c_{k,i}$ is the bias parameter of the convolution neural network. The output of convolution is downsampled:
\begin{equation}
w_i^{(k)}(m,n) = v_i^{(k)}(s_km,s_kn),
\label{downsampling}
\end{equation}
where $s_k$ is the downsampling factor. Finally, the downsampled signals processed by a generalized divisive normalization (GDN) transform:
\begin{equation}
u_i^{(k+1)}(m,n)=\frac{w_i{(k)}(m,n)}{(\beta_{k,i}+\sum_j\gamma_{k,ij}(w_j^{(k)}(m,n))^2)^{\frac{1}{2}}}.
\label{GDN}
\end{equation}
where $\beta_{k,i}$ and $\gamma_{k,ij}$ are the bias and scale parameters for the normalization operation.

Since the synthesis transform is the inverse operation of the analysis transform, all the parameters across all three stages, $\{h,c,\beta, \gamma\}$ will be optimized according to rate-distortion objective function in an end-to-end style. To deal with the zero derivatives due to the quantization, Ball\'{e} \emph{et al.} utilized an additive i.i.d uniform noise to simulate the quantizer in CNN training procedure, which enables the stochastic gradient descent approach to the optimization problem. This method outperforms JPEG2000 according to both PSNR and MS-SSIM metrics.
In addition, Ball\'{e} and his colleagues extended such model using the scale hyper priors for entropy estimation~\cite{balle2018variational}, which achieved similar objective coding performance with HEVC. Minnen \emph{et al.} continued to enhance the context model of entropy coding for end-to-end optimized image compression~\cite{minnen2018joint} and outperformed the HEVC intra coding. For future practical utility, both hardware-end support and the energy-efficiency analysis should be further explored since the autoregressive component is not easily parallelizable.
The image compression performance is further improved by Zhou \textit{et al.} by utilizing pyramidal feature fusion structure at the encoder and a CNN based post-processing filter at the decoder \cite{zhou2018variational}. The other end-to-end image compression work joint with quantization and entropy coding can be referred in \cite{agustsson2017soft,theis2017lossy}, and the CNN prediction based image compression can be can be referred in \cite{ahanonu2018lossless}.

\subsection{Recurrent Neural Network based Coding}

Unlike the CNN architecture mentioned above, RNN is a class of neural network with memory to store the recent behaviors. In particular, memory units in RNN have the connections to themselves, which transmit transformed information from the execution in the past. By taking advantage of these stored information, RNN changes the behavior of the current forward process to adapt to the context of current input. Hochreiter \emph{et al.} proposed the Long Short-Term Memory (LSTM) \cite{hochreiter1997long} to overcome the insufficiency of the decayed error backflow. More advanced improvements such as Gated Recurrent Unit (GRU) \cite{cho2014properties} are presented to simplify the recurrent evolution processes, and meanwhile they maintain the performance of the recurrent network in relevant tasks~\cite{chung2014empirical}. In analogous to CNN, for image compression task, RNN still suffers from the difficulties to propagate the gradients of the rate estimation.

\begin{figure}[t]
\center{
\includegraphics[width=0.45\textwidth]{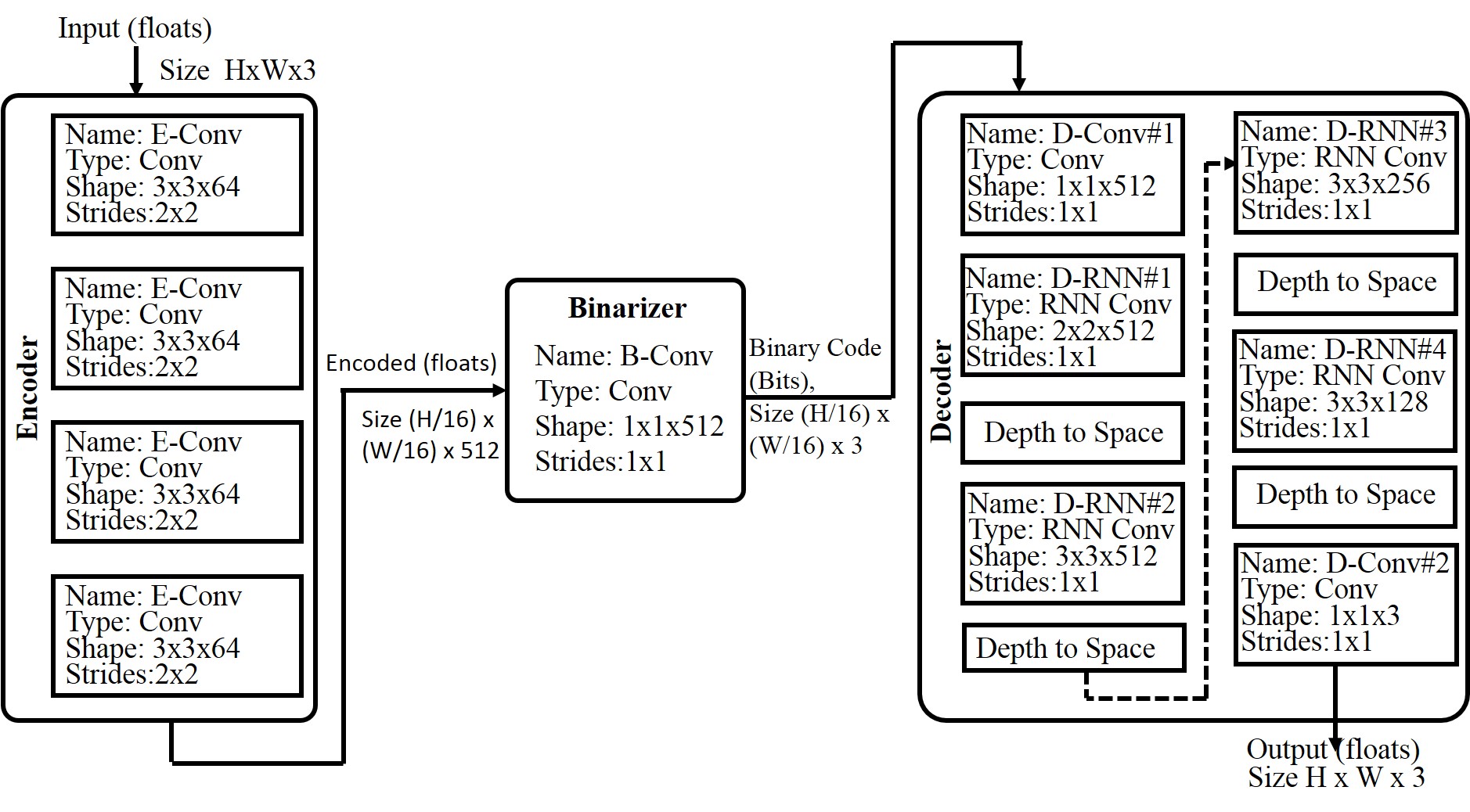}}
\caption{A single iteration of our shared RNN architecture \cite{toderici2017full}.}
\label{Fig:RNNImage}
\end{figure}

Toderici \emph{et al.} firstly proposed a RNN-based image compression scheme \cite{toderici2017full} by utilizing a scaled-additive coding framework to restrict the number of coding bits instead of the approximation of rate estimation in CNN \cite{balle2016end}. More specifically, the proposed method in \cite{toderici2017full} is an multi-iteration compression architecture supporting variational bitrate compression in progressive style. As shown in Fig.~\ref{Fig:RNNImage}, there are three modules in a single iteration, i.e., an encoding network \emph{E}, a binarizer \emph{B} and a decoding network \emph{D}, where \emph{D} and \emph{E} contain recurrent network components. The residual signals between the input image patch and the reconstructed one from decoding network \emph{D} can be further compressed into the next iteration.
To further improve the RNN-based image compression, Minnen \emph{et al.} presented a spatially adaptive image compression framework \cite{minnen2018spatially}. In this framework, the input images is divided into tiles which is similar to the existing image codecs such as the JPEG and JPEG2000. For each tile, an initial prediction is generated by a fully-convolutional neural network from the spatial neighboring tiles which have been decoded in the left and above regions. However, based on the released results, the proposed method only outperforms JPEG while it is inferior to JPEG2000.

\begin{figure}[t]
\center{
\includegraphics[width=0.45\textwidth]{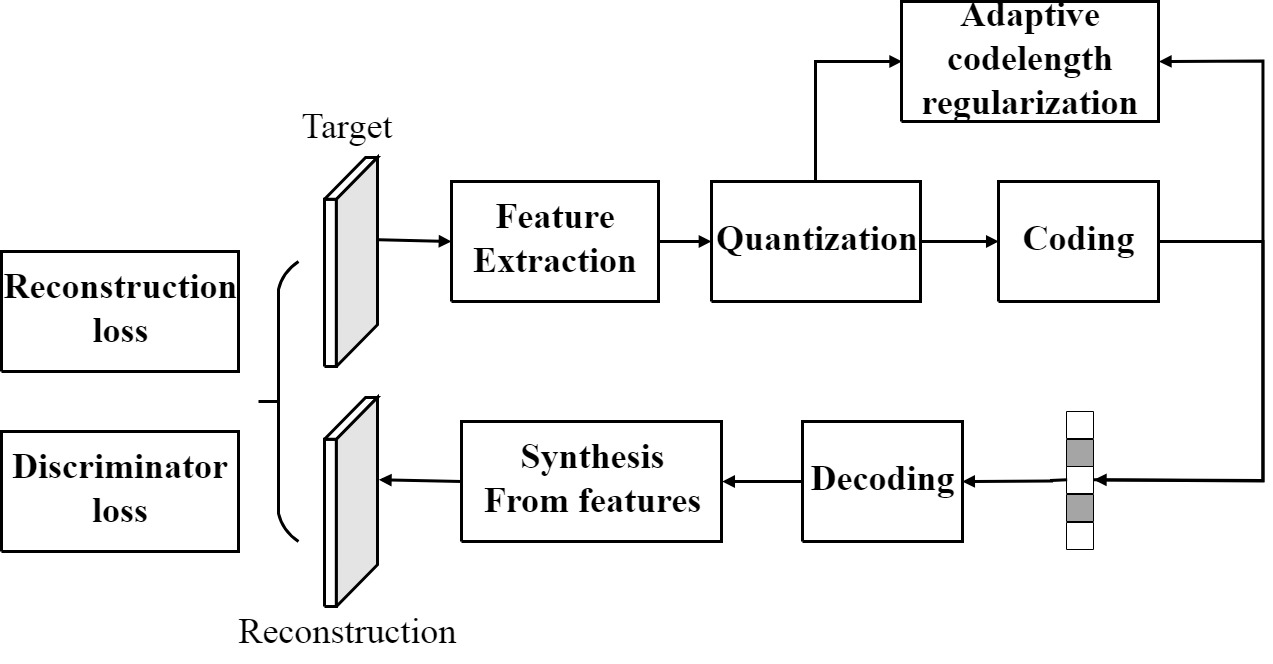}}
\caption{The overall architecture of GAN based image compression \cite{rippel2017real}.}
\label{Fig:GANImage}
\end{figure}

\subsection{Generative Adversarial Network based Coding}
Generative Adversarial Network is one of most attractive improvements in the application of deep neural network. GAN optimizes two network models i.e., generator and discriminator, simultaneously. Discriminator takes advantage of deep neural network to distinguish whether the samples are generated form the generator. At the same time, the generator is trained to overcome the discriminator and produce samples which pass the inspection. Adversarial loss has the advantage to assist the generator to improve the subjective quality of images and also can be designed for different tasks. In the image compression task, some research works focused on the perceptual quality of the decoded images and utilized GAN to improve the coding performance.

One of the representative works is proposed by Rippel and Bourdev in 2017 \cite{rippel2017real}, and it is an integrated and well optimized GAN based image compression, which not only achieves amazing compression ratio improvement but also can run in real-time by leveraging the massive parallel computation cores of GPU. As shown in Fig.~\ref{Fig:GANImage}, the input image is compressed into very compact feature space by networks as its compressed form, and the generative network is utilized to reconstruct the decoded image from the features. The most obvious difference between the GAN based image compression and those of CNN or RNN based schemes is the introduction of the adversarial loss which enhances the subjective quality of reconstructed image significantly. The generative network and adversarial network are trained jointly to significantly enhance the performance of the generative model. The GAN based method in \cite{rippel2017real} achieves significant compression ratio improvement, e.g., producing compressed files 2.5 times smaller than JPEG and JPEG2000, 2 times smaller than WebP, and 1.7 times smaller than BPG on generic images across all quality levels. Herein, the quality is measured by MS-SSIM, while the method is still not efficient using PSNR metric. Inspired by the advances in GAN based view synthesis, the light field (LF) image compression could achieve significant coding gain with generating the missing views using the sampled context views in LF \cite{jia2018light}. In particular, the contents generated by GAN are more consistent with the semantics of the original content than the specific textures. Especially, when enlarging the reconstructed images, we can see the content difference in specific textures.

In addition, Gregor \emph{et al.} introduced a homogeneous deep generative convolutional model \emph{DRAW} \cite{gregor2015draw} to the image compression task. Different from previous works, Gregor \emph{et al.} aimed to conceptual compression by generating the image semantic information as possible \cite{gregor2016towards}. A GAN-based framework for extreme image compression, targeting bitrates below 0.1 bpp, is explored in detail, which allows for different degrees of content generation \cite{agustsson2018extreme}. At present, the GAN-based compression is successful in narrow-domain images such as faces, and still needs more research on establishing models for general natural images.


\section{Advancement of Video Coding with Neural Networks}\label{CNNVideo}
The study of deep learning based video coding by leveraging the state-of-the-art video coding standard HEVC has been an active area of research in recent years. Almost all the modules in HEVC have been explored and improved by incorporating various deep learning models. In this section, we will review the development of video coding works with deep learning models from the five main modules in HEVC, i.e., intra prediction, inter-prediction, quantization, entropy coding and loop filtering. Finally, we will introduce several novel video coding paradigms, which are different from hybrid video coding framework.


\subsection{Intra Prediction Techniques using Neural Networks}


Although many neural network based image compression methods have been proposed and can be regarded as intra-coding strategy for video compression, their performances only surpass JPEG and JPEG2000 and are inferior to HEVC intra coding obviously. This also shows the superiority of the hybrid video coding framework. Therefore, many researchers focuses on video coding performance improvement by integrating the neural network techniques into hybrid video coding framework, especially into the state-of-the-art HEVC framework. Cui \textit{et al.} proposed an intra-prediction convolutional neural network (IPCNN) to improve the intra prediction efficiency, which is the first work integrating CNN into HEVC intra prediction. In IPCNN, the current $8\times 8$ block is firstly predicted according to HEVC intra prediction mechanism, and the best prediction version of current block generated by mode decision as well as its three nearest neighboring reconstructed $8\times 8$ blocks as additional context, i.e., the left block, the upper block and the upper-left block, composes a $16\times 16$ block, which is utilized as the input of IPCNN. The residual learning approach is adopted and the output of IPCNN is the residual block by subtracting the original blocks from the input ones. Then, the refined intra-prediction for the current $8\times 8$ block can be derived by subtracting the output residual block from the input one. The designed IPCNN not only heritages the powerful prediction efficiency of CNN, but also takes advantage of the far-distance structure information in spatial neighboring $8\times 8$ blocks instead of only utilizing one column plus one row reconstructed neighboring pixels as HEVC intra prediction. The additional context for prediction as well as the residue learning approach offers extra coding efficiency.

\begin{figure}[t]
\center{
\includegraphics[width=0.45\textwidth]{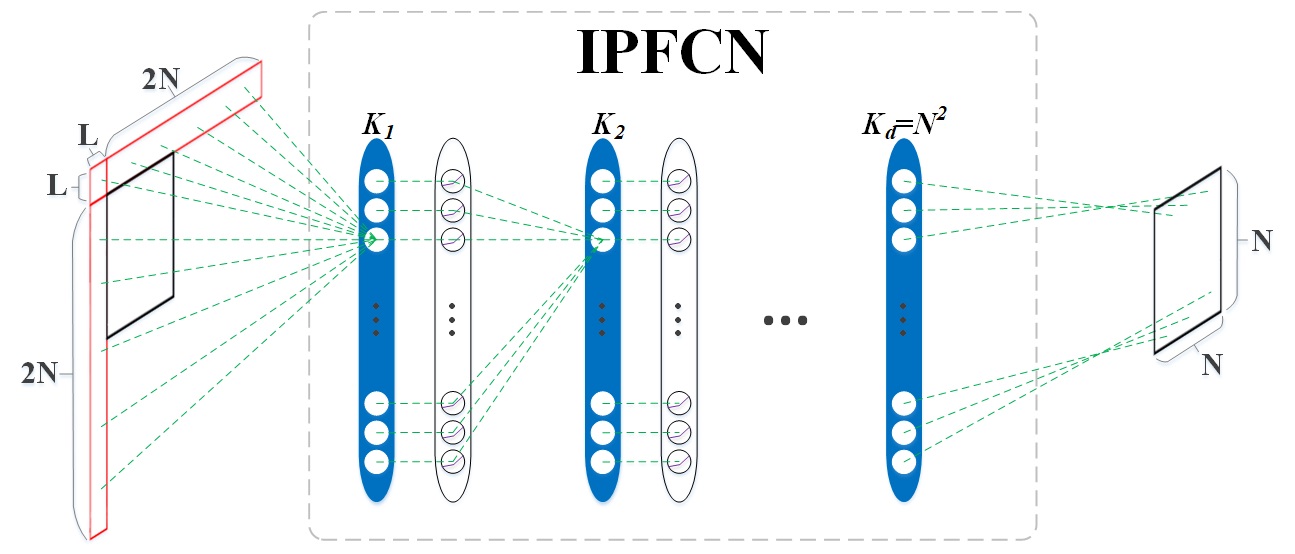}}
\footnotesize
\caption{The network structure of IPFCN \cite{li2018fully}.}
\label{Fig:IPFCN}
\end{figure}

Instead of using CNN to improve the quality of best HEVC intra prediction, Li~\textit{et al.} proposed a new intra prediction mode using fully connected network (IPFCN)~\cite{li2018fully}, which competes with the existing 35 HEVC intra prediction modes. Similar with IPCNN, IPFCN also utilizes neighboring multiple reference lines of reconstructed pixels as contextual input, but the prediction version of the current block from HEVC intra prediction is not utilized. Fig.\ref{Fig:IPFCN} shows the IPFCN structure, which is an end-to-end intra prediction mapping from reconstructed neighboring pixels to current block. Except for the output layer, each connected layer is followed by a nonlinear activation layer, where the parametric rectified linear unit (PReLU) is utilized. Each node of the output layer corresponds to a pixel. The corresponding coding performance as well as complexity is depicted in Table.~\ref{tab:IPFCN-HEVC}, where the abbreviation ``L" means light (which means parameter reduction version of models w/o ``L"), ``D" means dual (which means train one particular IPFCN model for DC and Planar modes, and another IPFCN model for the remaining angular modes), and ``S'' means single model (which means to train one model for all the intra prediction modes). The running time is tested on CPU platform. Compared with HEVC reference software HM-16.9, the proposed method achieved obvious bitrate saving, up to 3.0\% BD-rate saving on average. In particular, the IPFCN performs better for ultra high resolution 4K videos in class A, achieving up to 4.4\% BD-rate saving. However, the complexity is extremely high due to the fully connected neural nets and the float-point operations during the multiplication calculation, and there are up to more than 200 times increase for decoding as shown in Table \ref{tab:IPFCN-HEVC}. The CNN based chroma intra prediction is proposed in \cite{li2018hybrid} by utilizing both the reconstructed luma block and neighboring chrom blocks to improve intra chroma prediction efficiency. In \cite{Pfaff2018IntraJVET}, Pfaff \textit{et al.} proposed a more high efficiency intra prediction network under JEM software, and the running time of its simplification version only increase by 74\% and 38\% for intra encoding and decoding process with about 2.26\% BD-rate saving.


\begin{table}[t]
  \centering
  \renewcommand{\arraystretch}{1.15}\footnotesize
  	\caption{The coding performance of IPFCN~\cite{li2018fully} under common test condition with full length sequence.}
	\begin{tabular}{c|c|c|c|c}
		\hline
		\multirow{2}{*}{Sequences} & \multicolumn{4}{c}{IPFCN vs. HM-16.9}   \\ \cline{2-5}
		                           & IPFCN-S       & IPFCN-D        & IPFCN-S-L  & IPFCN-D-L       \\ \hline
		Class A                    & -3.8 \%       & -4.4 \%        & -3.0\%     & -3.7\%  \\ \hline
		Class B                    & -2.8 \%       & -3.2 \%        & -2.2\%     & -2.8\% \\ \hline
		Class C                    & -1.9 \%       & -2.1 \%        & -1.6\%     & -1.9\% \\ \hline
		Class D                    & -1.7 \%       & -1.8 \%        & -1.4\%     & -1.7\% \\ \hline
		Class E                    & -3.9 \%       & -4.5 \%        & -3.0\%     & -3.5\%  \\ \hline
		Overall                    & -2.6 \%       & -3.0 \%        & -2.1\%     & -2.5\% \\ \hline
        Encode Time                &4930\%         &13052\%   &285\%   &483\%    \\ \hline
        Decode Time                &26572\%        &28927\%   &923\%  &1141\%    \\ \hline
  \end{tabular}
  \label{tab:IPFCN-HEVC}
\end{table}

\begin{figure}[t]
\center{
\includegraphics[width=0.49\textwidth]{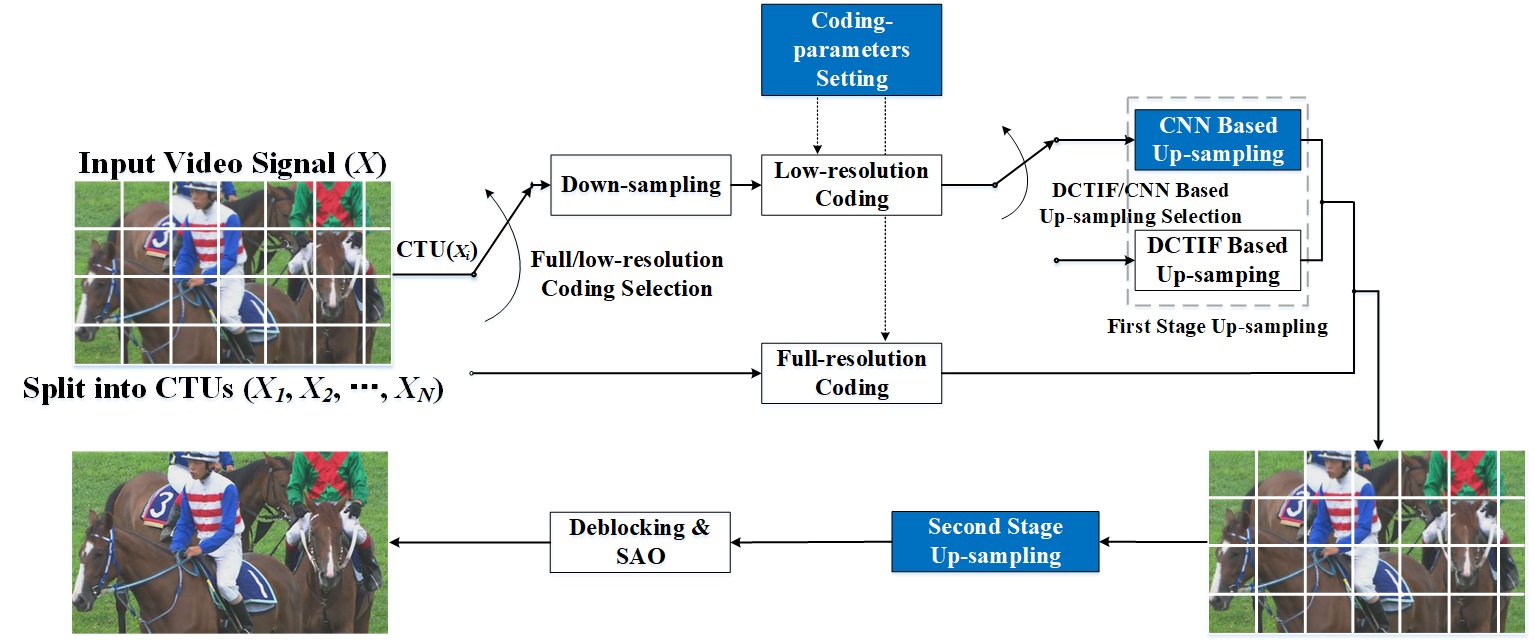}}
\footnotesize
\caption{The framework of neural network based intra prediction using up-sampling \cite{li2017convolutional}.}
\label{Fig:DUCNN}
\end{figure}

Instead of using neighboring reference samples to obtain block prediction, Li~\textit{et al.} explored CNN based down/up-sampling techniques as a new intra prediction mode for HEVC~\cite{li2017convolutional} and its extension for inter frame is proposed in \cite{lin2018convolutional}. Different from previous image-level down/upsampling techniques \cite{molina2006toward,shen2011down}, Li~\textit{et al.} designed the down/upsampling method in CTU-level and the framework is shown in Fig.~\ref{Fig:DUCNN}. In down/up-sampling mode, each CTU is firstly down-sampled into low resolution version, which is then coded using HEVC intra coding method. The upsampling is applied for the reconstructed low resolution CTU to restore its original resolution. To remove the boundary artifacts, a second stage upsampling CNN network is applied when the whole frame has been reconstructed. Then, the second-stage upsampling CNN can access to all the surrounding blocks of down/upsampling CTUs. To ensure the coding efficiency, a flag is signaled into bitstream to indicate whether the down/upsampling is switched on. Herein, the flag is determined according to the rate distortion optimization at the encoder. Due to the high efficiency of CNN based upsampling techniques, this work achieves significant coding gain especially at low bitrate scenario, around 5.5\% bitrate saving on average compared with HEVC. However, due to the limitations of the up-sampling algorithm, the bitrate saving for QPs (=22, 27, 32, 37) utilized in common test condition of HEVC is only 0.7\% for luma component.

In addition, the performance is also affected by the QPs used in compressed training video sequences, and the performance will degenerate when the test QPs deviate from those in the training stage. Based on the results using cross QP models, for the learned model from videos compressed at QP=$QP_0$, the performance degeneration when applying it to videos compressed at $QP_0+2$ is less than that of applying to videos compressed at $QP_0-2$. These results shows that the down/upsampling CNN prediction model is robust to the videos compressed by higher QPs. Beside intra prediction, Pfaff \textit{et al.} utilized a fully connected neural network with one hidden layer and neighboring reconstructed samples to predict the intra mode probabilities \cite{pfaff2018neural}, which can benefit the entropy coding module.

To alleviate the affects of compression noise on upsampling CNN, we proposed a dual-network based super-resolution strategy by bridging the low-resolution image and upsampling network using an enhancement network \cite{feng2018dual}. The enhancement network focuses on compressed noise reduction and feeds high quality input into the upsampling network. Compared with single upsampling network, the proposed method further improve coding performance at low bitrate scenario especially for ultra high resolution videos. In 2019, Li \textit{et al.} designed a compact representation CNN model to further improve the super-resolution CNN based compression framework by constraining the information loss of the low resolution images \cite{li2019learning}. Other CNN based intra coding techniques can be referred to \cite{zhang2017efficient,Hu2018Enhanced}, wherein the CTU level CNN enhancement model for intra coding is introduced in \cite{zhang2017efficient} and RNN based intra prediction using neighboring reconstructed samples is introduced in \cite{Hu2018Enhanced}.


\subsection{Neural Network based Inter Prediction}
In hybrid video coding, the inter prediction is realized by motion estimation on previous coded frames against the current frame, and in HEVC the precision of motion estimation is up to quarter-pixel, the value of which is calculated via interpolation, e.g., discrete cosine transform based interpolation filter (DCTIF)\cite{lv2012comparison}. Intuitively, the more similar of the inter predicted block and the current block are, the higher coding performance is achieved due to fewer prediction residuals left. Huo \textit{et al.} proposed a straightforward method \cite{huo2018convolutional} to improve inter prediction efficiency by utilizing the existing variable-filter-size residue-learning CNN (VRCNN) \cite{dai2017convolutional}, which is named CNN-based motion compensation refinement (CNNMCR). The CNNMCR jointly employs the motion compensated prediction and its neighboring reconstructed blocks as input of VRCNN, which is trained by minimizing the mean square errors between the input and its corresponding original signal. In fact, the improvement of CNNMCR for inter prediction is because the designed network can improve the inter prediction quality by reducing the compression noise and the boundary artifacts due to independent block processing.

\begin{figure}[t]
\includegraphics[width=0.40\textwidth]{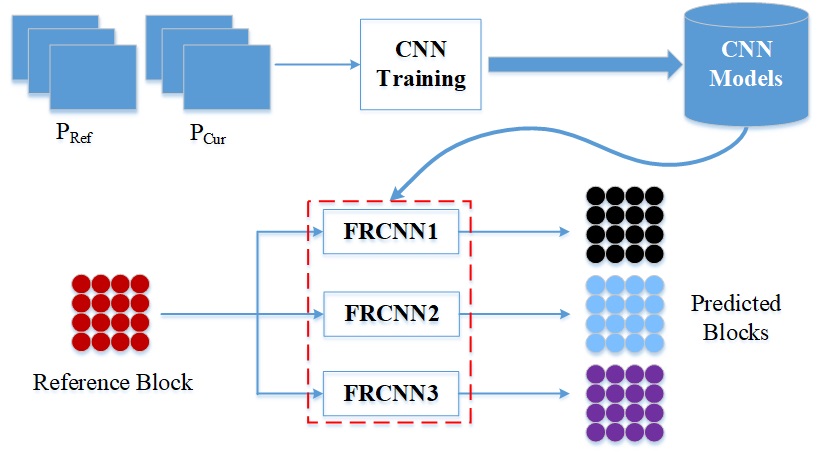}
\centering
    \caption{Framework of the FRCNN in~\cite{yan2018convolutional}.}
    \label{fig:FRCNN}
\end{figure}

Considering the importance of fractional-pixel motion compensation in inter prediction, Yan \textit{et al.} proposed a Fractional-pixel Reference generation CNN (FRCNN) to predict the fractional pixels \cite{yan2018convolutional}. This work is different from the previous interpolation or super-resolution problems, which predict pixel values in high resolution image, while FRCNN is to generate the fractional-pixels from reference frame to approach the current coding frame. Therefore, the fractional-pixel generation is formulated as a regression problem with the loss function as,
\begin{equation}
    f^\ast \equiv \mathop{\arg\min}_{f\in \bigtriangleup} \mathcal{L}(f(X),Y),
\label{FRCNNLossFunction}
\end{equation}
where $X$ is the motion compensation block by integer motion vector, $Y$ is current coding block, and $f$ is the regressor, which is implemented by CNN. Since the optimal position indicated by motion vectors may be in different fractional-pixel positions, e.g., three positions for half-pixel precision (0,1/2), (1/2,0) and (1/2,1/2). Thus, an individual CNN is trained for each fractional-pixel positions. Fig.\ref{fig:FRCNN} shows a training example for three half-pixel CNN models. In essence, the principle of FRCNN is the same with that of adaptive interpolation filters \cite{vatis2009adaptive}, the parameters of which are derived by minimizing the prediction errors at fractional-pixel positions and need to be transmitted to decoder side. The performance of FRCNN mainly thanks to the high prediction efficiency of CNN, and it achieves on average 3.9\%, 2.7\% and 1.3\% bitrate saving compared to HM-16.7, under Low-Delay P (LDP), Low-Delay B (LDB) and Random-Access (RA) configurations, respectively. However, the performance improvement comes from up to 120 FRCNN models for different slice types and 4 common QPs, which are trained from the specific videos compressed by HEVC under 4 common QPs and various coding configurations. Due to the poor generalization of the CNN models, the performance of FRCNN model may degenerate when applying it to the videos compressed by different configurations and QPs from training data, which is a potential problem to be solved in future. 

\begin{figure}[t]
\includegraphics[width=0.45\textwidth]{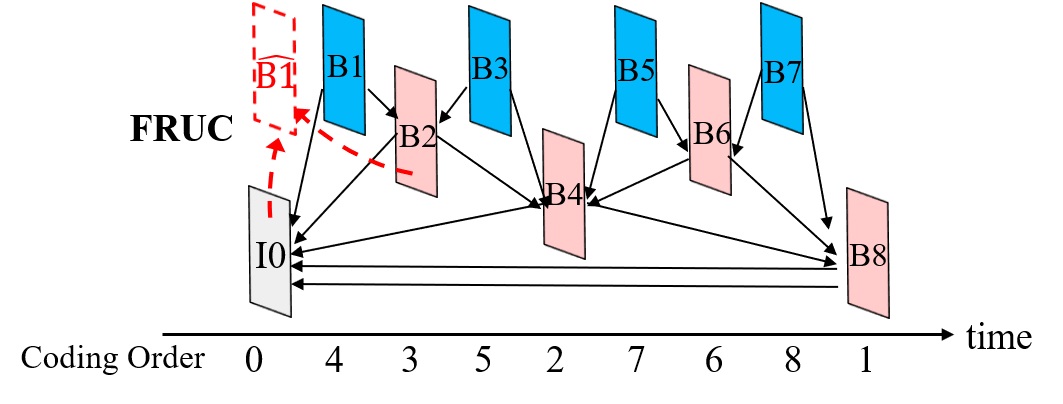}
\centering
\caption{Illustration for the proposed DVRF mode in~\cite{zhao2018Enhanced}.}
\label{fig:FRUC}
\end{figure}

Instead of improving the prediction performance for fractional-pixels, we directly explore the inter prediction block generation using CNN based frame rate up conversion (FRUC) techniques. A CNN based FRUC method in CTU level is proposed to generate a virtual reference frame $F_{Virtual}$, which is utilized as a new reference frame and named as direct virtual reference frame (DVRF) \cite{zhao2018Enhanced,Zhao2018EnhancedTIP}. As shown in Fig.~\ref{fig:FRUC}, current coding block can directly take the co-located block in $F_{Virtual}$ as inter prediction block without motion vectors. The state-of-the-art deep FRUC approach \textit{Adaptive Separable Convolution} \cite{niklaus2017video} is adopted and the two nearest bi-directional reference frames in the reference list are utilized as input for the network. This method achieves very promising compression performance, about 4.6\% bitrate saving compared with HM-16.9 and 0.7\% bitrate saving compared with JEM-7.1 \cite{Chen2015Algorithm} on average as shown in Table \ref{tab:BiPredFRUC}. Herein, JEM (Joint Exploration Model) is the reference software based on HEVC reference model for the JVET group, which is an organization working on the exploration of next generation video coding standard established by ITU-T Video Coding Experts Group (VCEG) and the ISO/IEC Moving Picture Experts Group (MPEG) in Oct. 2015. In addition, considering the limitation of the traditional bidirectional prediction using simple average of two prediction hypothesises, we further improve its efficiency by utilizing a six-layer CNN with $13\times 13$ receptive field size to infer the inter prediction block in a nonlinear fashion \cite{zhao2018cnn,Zhao2018EnhancedBi}, which achieves 2.7\% bitrate saving compared with HM-16.19 on average under the RA configuration as shown in Table \ref{tab:BiPredFRUC}. Although these methods obtained significant compression performance improvement, they also dramatically increase the run time for both encoding and decoding. The encoding and decoding time in Table \ref{tab:BiPredFRUC} are all tested with GPU for the convolution calculations in the proposed methods. The computation efficiency is still a severe problem for CNN based video compression techniques in practical applications. The CNN based fractional interpolation methods can be referred to \cite{zhang2017learning,yan2018convolutionalICIP,liu2019one}.

\begin{table}[t]
  \centering
  \renewcommand{\arraystretch}{1.15}\footnotesize
  	\caption{The coding performance of CNN-based bi-prediction and DVRF.}
	\begin{tabular}{c|c|c|c|c}
		\hline
		\multirow{2}{*}{Sequences} & \multicolumn{2}{c|}{BIP-CNN vs. HM-16.9}  &\multicolumn{2}{c}{DVRF}\\ \cline{2-5}
		                           & RA            & LDB  & RA (HM16.9)  & RA (JEM7.1)       \\ \hline
		Class A                    & -2.1 \%       & -1.7 \%    & -6.7\% & -1.3\%  \\ \hline
		Class B                    & -3.2 \%       & -1.9 \%    & -3.5\% & -0.4\% \\ \hline
		Class C                    & -2.2 \%       & -0.9 \%    & -4.0\% & -0.8\% \\ \hline
		Class D                    & -3.2 \%       & -1.0 \%    & -5.7\% & -0.7\% \\ \hline
		Class E                    & /             & -2.8 \%    & /      & -0.8\%  \\ \hline
		Overall                    & -2.7 \%       & -1.7 \%    & -4.6\% & -0.7\% \\ \hline
        Encode Time                &149\%          &185\%       &135\%   &124\%    \\ \hline
        Decode Time                &4259\%         &2853\%      &4376\%  &1025\%    \\ \hline
  \end{tabular}
  \label{tab:BiPredFRUC}
\end{table}

\subsection{Neural Network based Quantization and Entropy Coding for Video Coding}
In video coding, quantization and entropy coding are the lossy and lossless compression procedures respectively. For quantization part, the scalar quantization strategy has dominated hybrid video coding framework due to its low cost in computation and memory. However, this uniform scalar quantization does not conform to the characteristics of human visual system, and is not friendly to perceptual quality improvement. In \cite{alam2015perceptual}, Alam \textit{et al.} proposed a two-step quantization strategy using neural networks. In the first step, a CNN named VNet-2 is utilized to predict local visibility threshold $C_T$ for HEVC distortions of individual video frames, and the VNet-2 consists of 894 trainable parameters in three layers, i.e., convolution, subsampling and full connection, each of which contains one feature map. In the second stage, the quantization steps for $64\times 64$ CTU are derived by regression as,
\begin{equation}
\log(Q_{step}) = \alpha C_T^2 + \beta C_T + \gamma,
\end{equation}
where $C_T$ is predicted visibility threshold in the first stage, and $\{\alpha, \beta, \gamma\}$ are the model parameters related with patch features. The model parameters are predicted from three separate committees of neural networks respectively, and each committee had a total of five two-layered feed-forward networks with 10 neurons. Based on the proposed adaptive quantization strategy, on average 11\% bitrate saving can be obtained for luma channel against HEVC at the same perceptual quality measured by structural similarity (SSIM) \cite{wang2004image}.

After quantization, the syntax elements including coding modes and transform coefficients will be fed into entropy coding engine to further remove their statistical redundancy. HEVC adopts the CABAC as its entropy coding, which achieves very high coding efficiency mainly because there are many contexts designed to predict conditional probabilities accurately. Inspired by the prediction efficiency of CNN, Song \textit{et al.} improved the CABAC performance on compressing the syntax elements of 35 intra prediction modes by leveraging CNN to directly predict the probability distribution of intra modes instead of the handcrafted context models \cite{song2017neural}. The network architecture is based on LeNet-5 proposed by LeCun \textit{et al.} \cite{lecun1998gradient}, and the above-left, the above and the left reconstructed blocks with the same size of current coding block are utilized as one category of inputs. The other category inputs are the most probable modes (MPMs), each of which are transformed into a 35-dim one-hot binary vector. The output is a 35-dim vector recording the predicted probability values of the 35 modes. Due to the high prediction accuracy, the CNN based method can improve the CABAC performance achieving about 9.0\% bitrate saving for intra prediction mode coding when CU size is $8\times 8$. The similar principle can be applied to other syntax elements, e.g., motion vector, coefficients and transform indices. Puri \textit{et al.} applied the CNN to predict the optimal transform index probability distribution from the quantized coefficient blocks, and then utilized the probability to binarize the transform index using a variable length instead of a fixed length coding to improve the entropy coding performance \cite{puri2017cnn}. At present, the works on entropy coding are still limited and remain to be investigated, especially there are few CNN based work on the dominant syntax, quantized transform coefficients in video coding, which may bring more coding gains.


\subsection{Neural Network based Loop Filtering}
Loop filtering module is first introduced into video coding standard since H.263+ \cite{cote1998h}, and many different kinds of loop filters \cite{norkin2012hevc,fu2012sample,tsai2013adaptive,zhang2017low,ma2016nonlocal} are proposed after that. Especially, inspired by the success of CNN on image/video restoration filed, many of CNN based loop filters are designed to remove compression artifacts recently, which are much easier to implement the end-to-end training compared with other video coding modules. Zhang \textit{et al.} \cite{zhang2018residual} proposed a residual highway convolutional neural network (RHCNN) for loop filtering in HEVC. It is a deep network with 13 layers and the basic high way unit in each layer consists of two convolutional layers followed by the correponding activation function ReLUs and an identity skip connection. Since the compression noise levels are distinct for videos compressed with different QPs and frame types including I/B/P frames, the CNN models should be trained for different QP and frame type combinations, which lead to 156 CNN models for video coding application. To reduce memory cost for CNN based loop filters, Zhang \textit{et al.} merged the QPs into several bands, and trained the optimal RHCNNs for each band. Compared with HM-12.0, the RHCNN achieves about 5.7\%, 5.68\% and 4.35\% bitrate saving for I/P/B frames in low bit-rate circumstances respectively with 2$\sim$3 times of encoding time increase even using GPU and 20$\sim$30 times of encoding time increase using CPU.

By leveraging the coherence of the spatial and temporal adaptations, we improved the performance of CNN based loop filter, and designed the spatial-temporal residue network (STResNet) based loop filter \cite{jia2017spatial}. The loss function of STResNet is formulated as,
\begin{equation}
L(\Theta) = \frac{1}{N} \sum_{i=1}^N\|F(x_{i-1},x_i|\Theta)-y_i\|^2,
\end{equation}
where $(x_{i-1},x_{i},y_{i})$ are the training samples. $x_{i-1}$ and $x_{i}$ represent the $(i-1)^{th}$ and $i^{th}$ reconstructed frames and $y_i$ corresponds to the uncompressed $i^{th}$ frame. $F(x_{i-1},x_i|\Theta)$ represents the STResNet model, where $\Theta$ is the set of network parameters. Moreover, we further improved the filtering performance by introducing content-aware CNN based loop filter in \cite{Jia2018Content}. For a reconstructed frame, multiple CNN models are trained according to their filtering performance iteratively as that in \cite{zhang2017just}, and a corresponding discriminative network is also trained which is utilized to help select optimal filter in test stage to remove coding overheads. Compared with HEVC with/whitout ALF under HEVC common test condition (CTC), the proposed multi-model CNN filters achieve significant performance improvement as illustrated in Table \ref{tab:MMCNN} at the cost of explosive encoding and decoding run time increase even using GeForce GTX TITAN X GPU.

\begin{table}[t]
  \centering
  \renewcommand{\arraystretch}{1.15}\footnotesize
  \caption{The coding performance of Muti-Model CNN loop filter~\cite{Jia2018Content}. }
  \begin{tabular}{c|c|c|c|c|c|c}
  \hline
  \multicolumn{1}{c|}{\multirow{2}{*}{Sequences}}	&\multicolumn{3}{c|}{Anchor HM-16.9}	&\multicolumn{3}{c}{Anchor HM-16.9 + ALF} \\ \cline{2-7}
	        &AI        &LDB        &RA         &AI       &LDB      &RA           \\ \hline
  Class A	& -4.7\%   & -6.7\%    & -6.6\%	  & -2.7\%   & -3.2\%    & -3.1\%    \\ \hline
  Class B	& -3.5\%   & -5.7\%    & -6.5\%	  & -1.6\%   & -2.5\%    & -2.7\%    \\ \hline
  Class C	& -3.4\%   & -5.0\%    & -4.5\%	  & -3.4\%   & -4.0\%    & -3.7\%    \\ \hline
  Class D	& -3.2\%   & -3.8\%    & -3.3\%	  & -3.2\%   & -3.4\%    & -3.4\%    \\ \hline
  Class E	& -5.8\%   & -8.6\%    & -9.0\%	  & -4.3\%   & -5.8\%    & -5.3\%    \\ \hline
  Overall	& -4.1\%   & -6.0\%    & -6.0\%	  & -2.9\%   & -3.7\%    & -3.6\%    \\ \hline
  Encode Time &\multicolumn{3}{c|}{114\%}    &\multicolumn{3}{c}{108\%}      \\  \hline
  Decode Time &\multicolumn{3}{c|}{15010\%}  &\multicolumn{3}{c}{12800\%}        \\ \hline
  \end{tabular}
  \label{tab:MMCNN}
\end{table}

Although the CNN based loop filters have achieved substantial coding gains on the top of HEVC, these methods need to store multiple CNN models for different QPs, which increase the memory burdens for video codec. In \cite{song2018practical}, they provided an efficient solution for CNN based loop filters with memory efficiency. They combined QPs as an input fed into the CNN training stage by simply padding the scalar QPs into a matrix with the same size of input frames or patches. To some extent, this method alleviates the performance fluctuates of CNN based loop filters due to QP missing in training stage. Based on our experience, although the CNN based loop filters learned from combined QPs is a little inferior to QP-dependent CNN models, the performance loss is usually marginal. The residual prediction based CNN model of in-loop filter is proposed in \cite{park2016cnn} and the multi-scale CNN model for in-loop filter is designed in \cite{kang2017multi}.

Regarding the complexity of DL and none-DL based loop filtering methods under HEVC framework, the encoding time of~\cite{Jia2018Content} is 114\% and 108\% when the ALF is turned off/on respectively. However, the deocding time is drastically increased into 15010\% and 12800\% respectively. While the corresponding encoding and decoding complexity for ALF itself is 104\% and 123\%. Hence, there still exists large quantity of space and potential in optimization for DL based loop filtering algorithms in future studies, such as pruning and quantization for the float-point weights and biases in neural networks.

Besides the in-loop filters, there are also some post-filtering algorithms proposed to improve the quality of decoded video and images by reducing the compression artifacts. Dong \textit{et al.} proposed an end-to-end CNN \cite{dong2015compression} to remove the compression artifacts, which is learned in the supervised manner. The CNN architecture is work is derived from super-resolution network SRCNN \cite{dong2014learning} by embedding one or more ``feature enhancement'' layers after the first layer of SRCNN to clean the noisy features. Li \textit{et al.} proposed a universal model to deal with compressed image at different compression ratios \cite{li2017efficient} by utilizing a very deep CNN model. Yang \textit{et al.} proposed a multi-frame quality enhancement neural network for compressed video by utilizing the neighboring high quality frames to enhance the low quality frames. Herein, a support vector machine based detector is utilized to locate peak quality frames in compressed video \cite{yang2018multi}. CNN based quality enhancement also achieves convincing performance in the field of multiview plus depth video coding. Zhu \textit{et al.} designed CNN models for the post-processing of synthesized views to promote the 3D video coding performances \cite{zhu2018convolutional}. More works utilized more complicated structure to improve the compressed images \cite{cavigelli2017cas,zheng2018s}.


\subsection{New Video Coding Frameworks Based on Neural Network}
Although the elaborately designed hybrid video coding framework has achieved significant success on predominant compression performance, it becomes more and more difficult to be further improved. Moreover, it also becomes computation intensive and inhospitality to parallel computation as well as hardware manufacturer. Similar with neural network based image coding frameworks, some novel video coding frameworks are also investigated by assembling different neural network models. Chen \textit{et al.}  proposed a combination of several CNN networks called \emph{DeepCoder} which achieved similar perceptual quality with low-profiled x264 encoder~\cite{chen2017deepcoder}. In \emph{DeepCoder}, the intra prediction is implemented via a neural network to generate a feature map, denoted as \emph{fMap}, and the inter prediction is obtained from motion estimation on previous frames. The \emph{fMap} is further quantized and encoded into stream. The intra- and inter-prediction residuals are transformed into a more compact domain using neural networks, the process of which is similar with that of \emph{fMap} generation in intra prediction but with different neural network parameters. Both the \emph{fMaps} from intra prediction and residuals are quantized and coded using Huffman entropy coding. Although there are not as many coding tools as H.264/AVC, the \emph{DeepCoder} shows comparable compression performance compared with H.264/AVC, which shows a new solution for video coding.

Chen \textit{et al.} proposed a fully learning-based video coding framework by introducing the concept of VoxelCNN via exploring spatial-temporal coherence to effectively perform predictive coding inside learning network~\cite{chen2018learning}. Specifically, the proposed video coding framework can be divided into three modules, i.e., predictive coding, iterative analysis/synthesis and binarization. The VoxelCNN is designed to predict blocks in the video sequences conditioned on previously coded frames as well as the neighboring reconstructed blocks of current block. Then the compact discrete representation of the difference between predicted and original signals can be analyzed and synthesized in iterative manner using RNN model of Toderici \textit{et al.} \cite{toderici2017full}, which is composed of several LSTM-based auto-encoders with connections between adjacent stages. Finally, the bitstream is subsequently obtained after binarization and entropy coding. Although lack of entropy coding in their present work, the scheme still shows comparable performance with H.264/AVC, showing its potential in future video coding.

Inspired by the prediction for future frames of generative models \cite{ranzato2014video}, Srivastava \textit{et al.} proposed to utilize the Long Short Term Memory (LSTM) Encoder-Decoder framework to learn video representations in \cite{srivastava2015unsupervised}, which can be utilized to predict future video frames. There are mainly two models,  LSTM Autoencoder Model and LSTM Future Predictor Model, which consist of two recurrent neural networks. Different from Ranzato's work \cite{ranzato2014video} predicting one future frame, this model can predict a long future sequence into the future. Based on the experiments, with 16 input natural video frames, the model can reconstruct these 16 frames and predict the future 13 frames.

\section{Optimization Techniques for Image and Video Compression}\label{codingopt}

The state-of-the-art video coding standard, HEVC, achieves the optimal compression performance by exhaustively traverse all the possible coding modes and partitions to determine the optimal coding parameters according to rate-distortion costs. The computational costs can be extremely reduced by predicting the optimal coding parameters to skip unnecessary RD calculations. The fast mode-decision algorithms are proposed for coding unit (CU) and prediction unit (PU) respectively on basis of neural networks, which are not only parallel-friendly but also easy for VLSI design \cite{liu2016cu,song2017cnn}. More specifically, the fast algorithm first carries out a coarse analysis based on the local gradients to classify the blocks into homogeneous and edge categories. This strategy not only can reduce the burden of CNN but also can make CNN avoid ill-conditions due to homogeneous blocks. Then, the CNN is designed for edge blocks to decrease no less than two CU partition modes in each CTU for full rate-distortion optimization process. The designed network contains one convolution layer with one max pooling layer followed by three full connected layers and takes the QP values into network at the last fully connected layer. Each square CU is used as network input while the output is the binary decision of quad-split or no-split for current CU. As such, the recursive mode traverse and selection process is eliminated. On average, their method achieves 61.1\% intra coding time saving, whereas the BD rate loss is only 2.67\% compared with HM-12.0.
Xu \textit{et al.} predicted the entire CTU partition structure by using both CNN and LSTM to determine whether the mode decision should be early terminated~\cite{xu2018reducing}.

\section{Conclusions and Outlook}\label{conclusion}
Image and video compression aims to seek more compact representation for visual signals while keeping high quality, and become more and more important in big visual data era. In this paper, the neural network based image and video compression techniques have been reviewed, especially for the recent deep learning based image and video compression techniques.
With the survey presented earlier in this paper, it is apparent that the state-of-the-art neural network based end-to-end image compression is still in its infancy which only outperforms the JPEG2000 and struggles against HEVC.
The marriage of neural network and traditional hybrid video coding framework obtained significant performance improvement compared with the latest video coding standard, HEVC. This demonstrates the advantages of both neural networks and hybrid video coding framework.

Based on the review, we think that the advantages of neural network in image and video compression are three folds. First, the excellent content adaptivity of neural network is superior to signal processing based model because the network parameters are derived based on lots of practical data while the models in the state-of-the-art coding standards are handcrafted based on image and video prior knowledge. Second, the larger receptive field is widely utilized in neural network models which not only utilizes the neighboring information but also can improve coding efficiency by leveraging samples from far distance, but the traditional coding tools only utilized the neighboring samples and are difficult to utilize far distant samples. Third, the neural network can well represent both texture and feature, which makes the joint compression optimization for both human view and machine vision analysis. However, the existing coding standards only pursue high compression performance toward human view task.

We envision that deep learning based image/video compression will play more important roles in representing and delivering images and videos with better quality and fewer bitrates, and the following confronted issues are required to be further investigated:
\begin{itemize}
\item \textbf{Semantic-fidelity oriented image and video compression.} Along with the fast development of computer vision techniques and explosively increasing of images and videos, the visual signal receivers are not only human visual system, but also the computer vision algorithms. Meanwhile, the neural network especially deep learning techniques are more appropriate for sematic information representation based on its great success in image and video understanding tasks. Therefore, the sematic-fidelity will become critical for further applications as well as traditional visual-fidelity requirement.


\item \textbf{Rate-distortion (RD) optimization guided neural network training and adaptive switching for compression task.} The rate-distortion theory is the key of the success for traditional image and video compression, but it has not been well explored in current neural network based compression tasks. A single network to deal with all the images and videos with diverse structures is inefficient obviously. Therefore, the multi-network adaptively training and switching according to RD is a possible solution.

\item \textbf{Memory and computation efficient design for practical image and video codec.} The biggest obstacle in hindering the deployment of deep learning based image and video compression is the burdens in computation and memory. To achieve high performance, larger neural networks with more layers and nodes
    are usually considered, but the various efficiency of network parameters are not well explored. For image and video compression problem, at present, there is no related research work by jointly considering both the compression performance and the efficiency in computation and memory for neural networks, which is important for practical applications.
\end{itemize}

For the semantic-friendlily oriented image and video compression, we have attempted to design innovative visual signal representation framework to elegantly support both human vision viewing and machine vision analysis. In view of the lightweight and the importance of features for visual semantic descriptors, e.g., CNN features, we proposed the hierarchical visual signal representation in \cite{zhang2017joint} by jointly compressing the feature descriptors and visual content. More specially, for each video frame, feature descriptors are first extracted and compressed, and then the decoded features are utilized to assist visual content compression by handling large-scale global motion. This strategy not only improves the visual content compression efficiency but also ensures the visual analysis performance due to the feature extraction from original video without the influence of compression artifacts. In \cite{Li2018Joint}, we further investigated the novel visual signal representation structure with deep learning based end-to-end image compression framework, which can directly conduct more image understanding tasks from the compression domain. The rationale behind this approach lies in that the neural network architectures commonly used for learned compression (in particular the encoders) are similar to the ones commonly used for inference, and learned image encoders are hence, in principle, capable of extracting features relevant for inference tasks. As such, this approach could be extended in the future to simultaneously train and learn for the end-to-end image compression and understanding.


In CNN based image and video compression, the CNN model compression is also a multi-variable optimization problem, which should be optimized jointly considering computational cost, CNN performance and rates utilized for CNN transmission (if needed). The previous work \cite{su2009complexity} proposed a complexity-distortion optimization formulation under power constraints for video coding problem, which can be further extended to CNN model compression optimization jointly with computational costs and video compression performance.

Based on the discussion of this paper, neural network has also shown promising results on future image and video compression tasks. Although there are still many problems in computational complexity and memory consumption, their high efficiency in prediction and compact representation for image and video signals has made neural network obtain substantial coding gain on top of the state-of-the-art video coding frameworks. Their intrinsic parallel-friendly attribute also makes them suitable for the largely deployed parallel computation architectures, e.g., GPU and TPU. Moreover, the network based end-to-end optimization approaches are more flexible than hand-crafted methods, and they can be rapidly optimized or tuned, which also makes the network with enormous potential in further image and video compression problem as well as other artificial intelligence problems.





\ifCLASSOPTIONcaptionsoff
  \newpage
\fi



\bibliographystyle{IEEEtran}
\bibliography{IEEEabrv,DeepOverview-v3}
%



%
\begin{IEEEbiography}[{\includegraphics[width=1in,height=1.25in,clip,keepaspectratio]{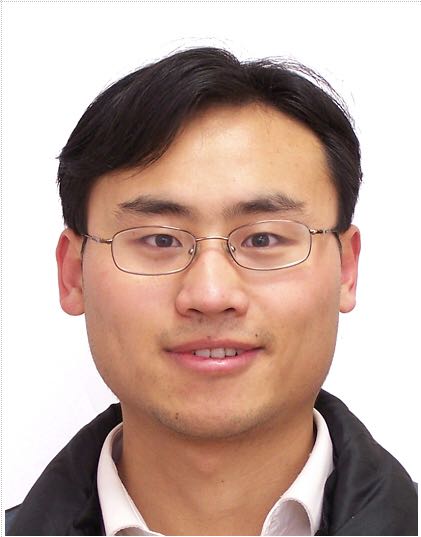}}]{Siwei Ma}
(M'03-SM'12) received the B.S. degree from Shandong Normal University, Jinan, China, in 1999, and the Ph.D. degree in computer science
from the Institute of Computing Technology, Chinese Academy of Sciences, Beijing, China, in 2005. He held a post-doctoral position with the University of Southern California, Los Angeles, CA, USA, from 2005 to 2007. He joined the School of Electronics Engineering and Computer
Science, Institute of Digital Media, Peking University, Beijing, where he is currently a Professor. He has authored over 200 technical articles in refereed journals and proceedings in image and video coding, video processing, video streaming, and transmission. He is an Associate Editor of the IEEE TRANSACTIONS ON CIRCUITS AND SYSTEMS FOR VIDEO TECHNOLOGY and the \textit{Journal of Visual Communication and Image Representation}.
\end{IEEEbiography}

\begin{IEEEbiography}[{\includegraphics[width=1in,height=1.25in,clip,keepaspectratio]{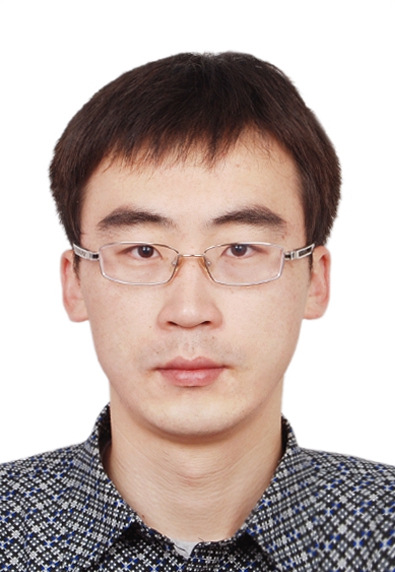}}]{Xinfeng Zhang}
(M'16) received the B.S. degree incomputer science from the Hebei University of Technology, Tianjin, China, in 2007, and the Ph.D. degreein computer science from the Institute of Computing Technology, Chinese Academy of Sciences, Beijing, China, in 2014. From 2014 to 2017,  he  was a Research Fellow with the Rapid-Rich Object Search Lab, Nanyang Technological  University, Singapore. From Oct. 2017 to Oct. 2018, he was a Post-Doctoral Fellow with  the School of Electrical Engineering System, University of Southern California, Los Angeles, CA, USA. From Dec. 2018 to Aug. 2019, he was a Research Fellow with the department of Computer Science, City University of Hong Kong. He currently is an Assistant Professor with the Department of Computer Science, University of Chinese Academy of Sciences. He has proposed over 20 technical proposals to ISO/MPEG, ITU-T and AVS standards, and authored more than 100 refereed journal/conference papers. He received the Best Paper Award of IEEE Multimedia 2018, the Best Paper Award at the 2017 Pacific-Rim Conference on Multimedia (PCM), and is the coauthor of a paper that received the Best Student Paper Award in IEEE International Conference on Image Processing 2018. His research interests include video compression, image/video quality assessment, and image/video analysis.
\end{IEEEbiography}

\begin{IEEEbiography}[{\includegraphics[width=1in,height=1.25in,clip,keepaspectratio]{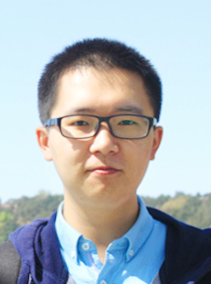}}]{Chuanmin Jia}
(S'18) received B.E. degree in computer science from Beijing University of Posts and Telecommunications, Beijing, China, in 2015.
He is currently pursuing Ph.D. degree with the Department of Computer Science in Peking University, Beijing, China.
He was a visiting student with Video Lab, New York University, NY, USA, in 2018.
His research interests include video compression, light field compression and machine learning.

He received the Best Paper Award of the Pacific-Rim Conference on Multimedia (PCM) in 2017, IEEE Multimedia in 2018 and Best Student Paper Award in IEEE International Conference on Multimedia Information Processing and Retrieval 2019.
\end{IEEEbiography}

\begin{IEEEbiography}[{\includegraphics[width=1in,height=1.25in,clip,keepaspectratio]{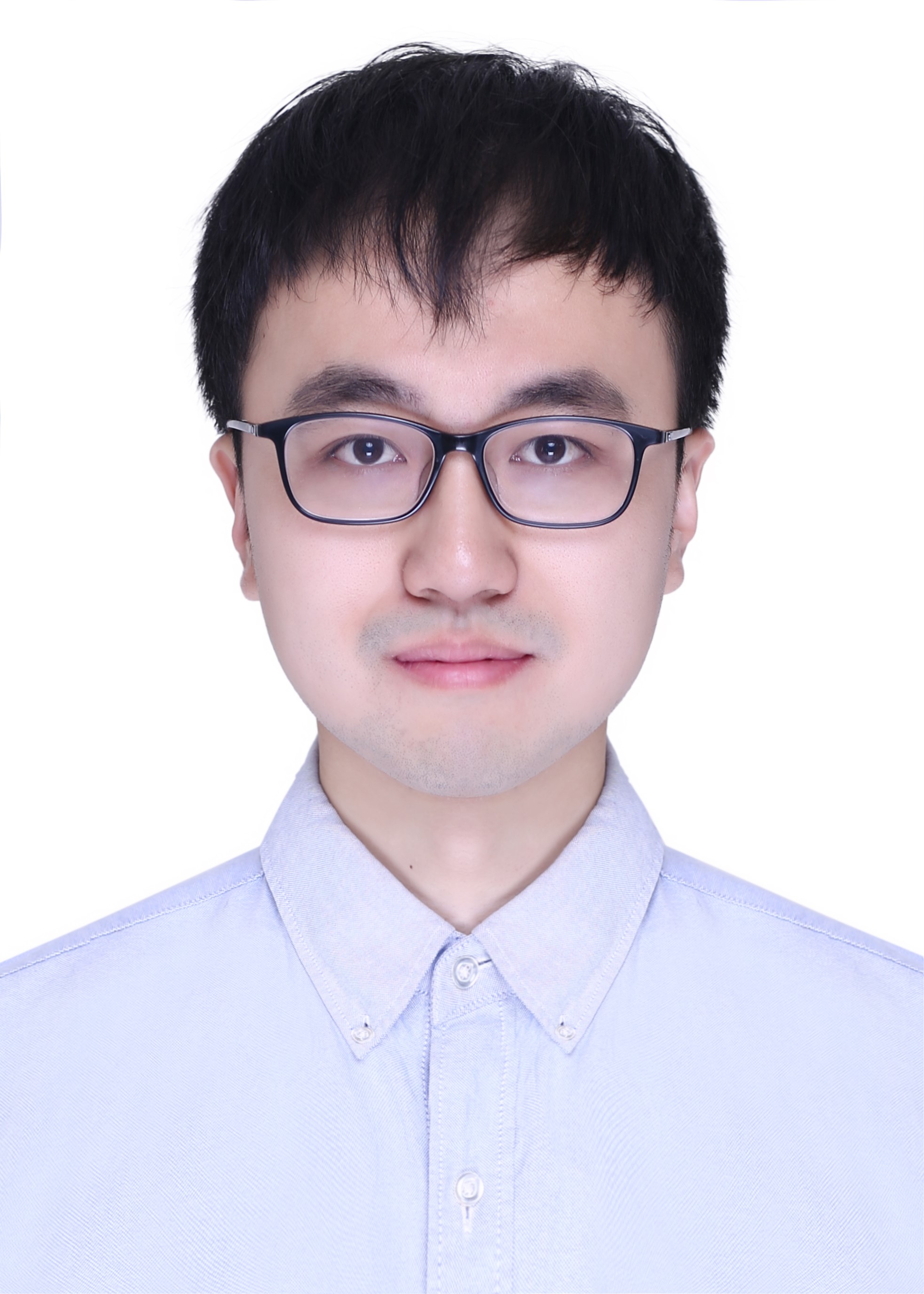}}]{Zhenghui Zhao} received the B.S. degree from School of Mathematical Sciences, Dalian University of Technology in 2015. He is currently working toward the Ph.D. degree at Peking University. His research interests include image and video compression, image and video processing.
\end{IEEEbiography}

\begin{IEEEbiography}[{\includegraphics[width=1in,height=1.25in,clip,keepaspectratio]{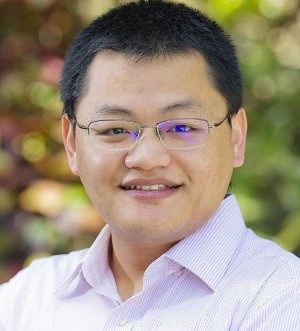}}]{Shiqi Wang} received the B.S. degree in computer science from the Harbin Institute of Technology in 2008, and the Ph.D. degree in computer application technology from the Peking University, in 2014. From Mar. 2014 to Mar. 2016, he was a Postdoc Fellow with the Department of Electrical and Computer Engineering, University of Waterloo, Waterloo, Canada. From Apr. 2016 to Apr. 2017, he was with the Rapid-Rich Object Search Laboratory, Nanyang Technological University, Singapore, as a Research Fellow. He is currently an Assistant Professor with the Department of Computer Science, City University of Hong Kong. He has proposed over 40 technical proposals to ISO/MPEG, ITU-T and AVS standards, and authored more than 150 refereed journal/conference papers. He received the Best Paper Award of IEEE Multimedia 2018, the Best Paper Award at the 2017 Pacific-Rim Conference on Multimedia (PCM), and is the coauthor of a paper that received the Best Student Paper Award in IEEE International Conference on Image Processing 2018. His research interests include video compression, image/video quality assessment, and image/video search and analysis.
\end{IEEEbiography}

\begin{IEEEbiography}[{\includegraphics[width=1in,height=1.25in,clip,keepaspectratio]{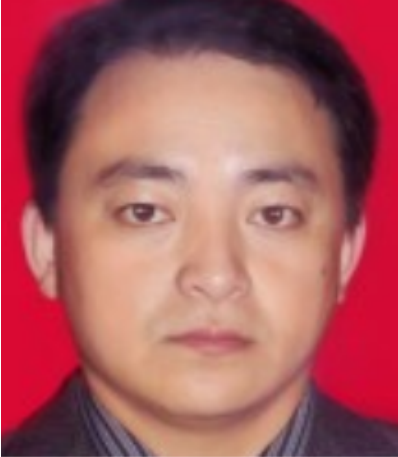}}]{Shanshe Wang}
Shanshe Wang received the B.S. degree from the Department of Mathematics, Heilongjiang University, Harbin, China, in 2004, the M.S. degree in computer software and theory from Northeast Petroleum University, Daqing, China, in 2010, and the Ph.D. degree in computer science from the Harbin Institute of Technology. He held a postdoctoral position with Peking University from 2016 to 2018. He joined the School of Electronics Engineering and Computer Science, Institute of Digital Media, Peking University, Beijing, where he is currently a Research Assistant Professor. His current research interests include video compression and image and video quality assessment.
\end{IEEEbiography}

%
%
%




\end{document}